\documentclass[10pt,twocolumn,letterpaper]{article}

\usepackage[pagenumbers]{cvpr} 

\usepackage{gensymb,soul}
\usepackage{graphics} 
\usepackage{booktabs}
\usepackage{epsfig} 
\usepackage{cite}
\usepackage{pifont}
\usepackage{mathptmx} 
\usepackage{amsmath} 
\usepackage{amssymb}  
\usepackage[table,xcdraw]{xcolor}
\usepackage[linesnumbered,ruled]{algorithm2e}
\usepackage{algpseudocode}
\usepackage{multirow}

\newcommand{\mytabref}[1]{Table~\ref{#1}}
\newcommand{\myeqref}[1]{(\ref{#1})}
\newcommand{\myfigref}[1]{Figure~\ref{#1}}
\newcommand{\myalgref}[1]{Algorithm~\ref{#1}}
\newcommand{\mysecref}[1]{Section~\ref{#1}}

\newcommand{\colorfirst}{\cellcolor[HTML]{c0e2ca}}
\newcommand{\colorsecond}{\cellcolor[HTML]{fff5b3}}
\newcommand{\colorthird}{\cellcolor[HTML]{ffd9b3}}

\algnewcommand\algorithmicswitch{\textbf{switch}}
\algnewcommand\algorithmiccase{\textbf{case}}
\algnewcommand\algorithmicassert{\texttt{assert}}
\algnewcommand\Assert[1]{\State \algorithmicassert(#1)}%
\algdef{SE}[SWITCH]{Switch}{EndSwitch}[1]{\algorithmicswitch\ #1\ \algorithmicdo}{\algorithmicend\ \algorithmicswitch}%
\algdef{SE}[CASE]{Case}{EndCase}[1]{\algorithmiccase\ #1}{\algorithmicend\ \algorithmiccase}%
\algtext*{EndSwitch}%
\algtext*{EndCase}%

\SetAlgoSkip{}

\DeclareMathAlphabet{\mathcal}{OMS}{cmsy}{m}{n}
\definecolor{cvprblue}{rgb}{0.21,0.49,0.74}
\usepackage[pagebackref,breaklinks,colorlinks,citecolor=cvprblue]{hyperref}

\def\paperID{1234} 
\def\confName{CVPR}
\def\confYear{2025}

\title{GS-LIVM: Real-Time Photo-Realistic LiDAR-Inertial-Visual Mapping with Gaussian Splatting}

\author{
Yusen Xie$^{1}$ \qquad  Zhenmin Huang$^{2}$ \qquad  Jin Wu$^{2}$ \qquad Jun Ma$^{1,2}$\\
$^{1}$Robotics and Autonomous Systems Thrust\\The Hong Kong University of Science and Technology (Guangzhou) \\  $^{2}$Department of Electronic and Computer Engineering\\The Hong Kong University of Science and Technology \\
\tt\small \{yxie827@connect.hkust-gz.edu.cn; zhuangdf@connect.ust.hk;\\ \tt\small  jwucp@connect.ust.hk; jun.ma@ust.hk\}
\\
\tt\small \url{
https://github.com/xieyuser/GS-LIVM
}
}

\begin{document}
\twocolumn[{
\maketitle
 \begin{center}
 \centering
 \vspace{-0.3in}
 \includegraphics[width=0.99\linewidth]{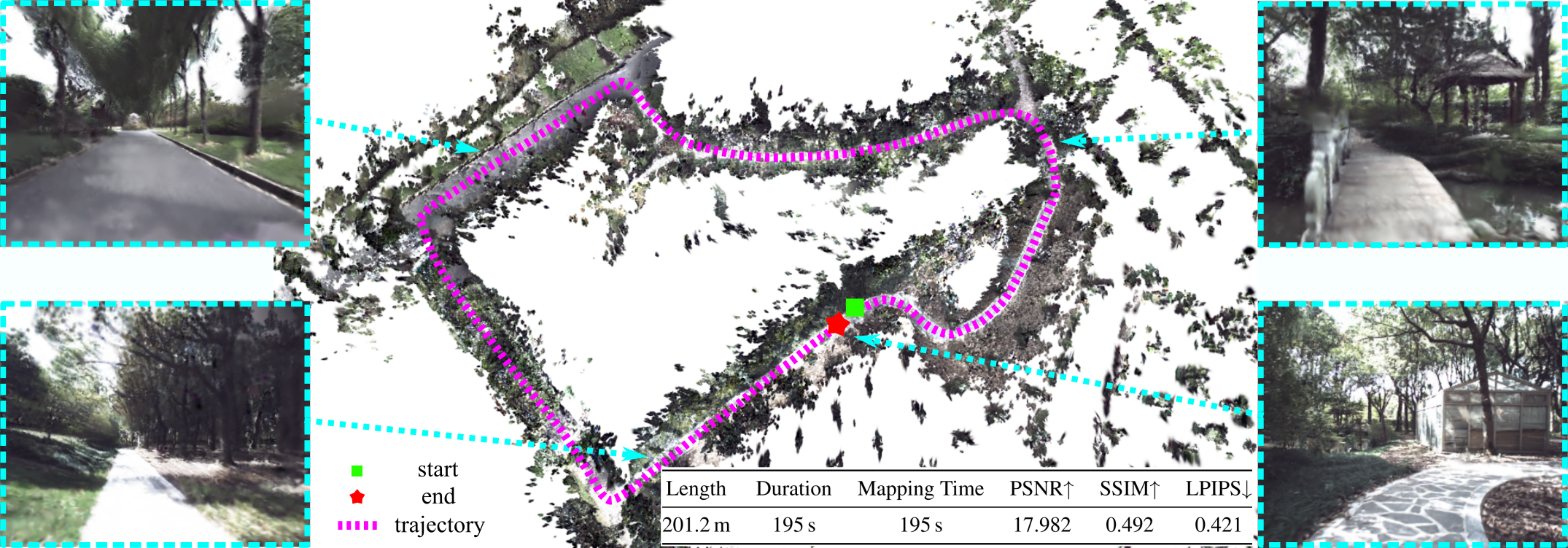}
     \captionof{figure}{Illustration of our photo-realistic mapping results on Botanic Garden Sequence \textit{1018\_13}. The pink dashed line represents the running trajectory, with the entire trajectory being approximately 200 meters. The green square is the starting point and the red pentagram is the endpoint. The rendering images at different locations are shown in the corners. The quantitative metrics are shown in the table.}
     \label{fig:firstdemo}
 \end{center}
}]


\begin{abstract}
In this paper, we introduce \textbf{GS-LIVM}, a real-time photo-realistic LiDAR-Inertial-Visual mapping framework with Gaussian Splatting tailored for outdoor scenes.
Compared to existing methods based on Neural Radiance Fields (NeRF) and 3D Gaussian Splatting (3DGS), our approach enables real-time photo-realistic mapping while ensuring high-quality image rendering in large-scale unbounded outdoor environments. 
In this work, Gaussian Process Regression (GPR) is employed to mitigate the issues resulting from sparse and unevenly distributed LiDAR observations. 
The voxel-based 3D Gaussians map representation facilitates real-time dense mapping in large outdoor environments with acceleration governed by custom CUDA kernels.  
Moreover, the overall framework is designed in a covariance-centered manner, where the estimated covariance is used to initialize the scale and rotation of 3D Gaussians, as well as update the parameters of the GPR.
We evaluate our algorithm on several outdoor datasets, and the results demonstrate that our method achieves state-of-the-art performance in terms of mapping efficiency and rendering quality. The source code is available on GitHub
.
\end{abstract}    
\section{Introduction}
\label{sec:intro}

Over the past two decades, simultaneous localization and mapping (SLAM) has emerged as a cornerstone technology underpinning advancements in robotics~\cite{hess2016real} and autonomous driving~\cite{Geiger2012CVPR, geiger2013vision}.
Typically, it relies on sensor fusion techniques that leverage sparse feature representations~\cite{fastlivo, fastlio, fastlio2, r2live, r3live, yuan2024sr}. While effective in various contexts, these approaches have shown limitations in achieving high-quality performance in 3D reconstruction and photo-realistic environmental rendering.
The landscape of research in SLAM begins to shift with the advent of powerful neural scene representation techniques, notably Neural Radiance Fields (NeRF)~\cite{Mildenhall20eccv_nerf, rosinolnerf2023} and 3D Gaussian Splatting (3DGS)~\cite{kerbl20233d,keetha2024splatam, kerbl2024hierarchical}. This paradigm shift leads to the development of photo-realistic neural SLAM methods, which greatly improve the fidelity of novel view-synthesis (NVS) and environmental representation.


Despite these advancements, it leaves an open problem in the application of these novel scene representation techniques to large-scale unbounded outdoor scenes. Current methods have demonstrated remarkable performance in indoor settings utilizing RGB-D sensors~\cite{yan2024gs, keetha2024splatam, matsuki2024gaussian, yugay2023gaussian, huang2024photo, sucar2021imap} and monocular cameras~\cite{matsuki2024gaussian, huang2024photo}. However, their applicability to outdoor scenes, where data acquisition is generally performed through multi-line spinning LiDAR and non-repetitive scanning LiDAR, is constrained. These constraints are further compounded by the uncertainty of point cloud scanning trajectories, leading to uneven point cloud distributions. Moreover, in outdoor SLAM scenarios, the movement is predominantly unidirectional, which leads to a growing bias towards the camera direction during the optimization of 3D Gaussians~\cite{kerbl20233d, matsuki2024gaussian}. This directional bias results in a significant degradation of rendering quality when observed from new viewpoints. Several studies~\cite{hong2024liv, lang2024gaussian,yan2024street,zhou2024drivinggaussian} have explored the integration of dense neural scene representations with conventional SLAM frameworks for outdoor scenes, yet significant challenges are encountered. The extensive optimization time required for offline mapping can result in over-fitting of 3D Gaussian representations to supervised information, yielding higher image evaluation metrics from these specific viewpoints~\cite{zhou2024drivinggaussian,yan2024street,hong2024liv}. These leads to both inefficient processing and compromised image synthesis quality from new perspectives, highlighting the complexities of developing effective and versatile outdoor SLAM systems.

With the aforementioned discussions as a backdrop, we propose \textbf{GS-LIVM} to achieve real-time performance on photo-realistic 3DGS reconstruction for large-scale unbounded outdoor scenes.
Our approach initiates with tightly-coupled LiDAR-Inertial-Visual odometry~\cite{r3live, yuan2024sr} for precise state estimation and coordinate transformation of colored point clouds.
To address the issues of sparsity and uneven distribution inherent in LiDAR point clouds, we apply Gaussian Process Regression (GPR)~\cite{ruan2020gp, seeger2004gaussian, li2020gp} to predict an evenly distributed point cloud as part of our photo-realistic mapping process.
The covariance output from the GPR module informs the update of GPR parameters and contributes to initializing scales and rotations of 3D Gaussian, which accelerates 3D Gaussian optimization process.
We further leverage LiDAR's new observations of the environment to iteratively update the parameters of GPR and the structure of 3D Gaussians. 
By continuously refining our framework with covariance estimates and image rendering data, we achieve high-quality 3D photo-realistic reconstruction and photo-realistic rendering in outdoor scenes. 
\myfigref{fig:firstdemo} shows the results of our real-time
photo-realistic mapping in Botanic Garden \textit{1018\_13} sequence.
To the best knowledge of the authors, our development is the first real-time photo-realistic SLAM framework tailored for LiDAR-Inertial-Visual fusion applications in large-scale unbounded outdoor scenes.

In this paper, our main contributions are summarized as follows: 
\begin{itemize}

\item We employ GPR within voxel-level (Voxel-GPR) to construct a uniform point cloud mesh grid, which effectively addresses sparse and unevenly distributed LiDAR observations. Voxel-GPR facilitates the utilization of fewer 3D Gaussians than the original input without compromising the performance, and this substantially reduces GPU memory consumption and enhances the speed of 3D Gaussians optimization.

\item  We develop a fast initialization technique for the scale and rotation parameters of the 3D Gaussians, which enables rapid convergence during map expansion. This approach alleviates the decline in rendering quality resulting from inadequate optimization iterations during fast movements of the sensor platform.


\item We devise an iterative optimization framework centered on variance, for iteratively reducing noise and hastening the convergence of the Voxel-GPR module.  Moreover, leveraging new observations from LiDAR, we introduce a structural similarity regularization term to the existing 3D Gaussian map, thus bolstering the robustness of NVS.


\item We conduct thorough validation on various outdoor SLAM datasets using various LiDAR, IMU, and camera configurations. The results showcase the superior capability of our algorithm to achieve real-time photo-realistic mapping in large-scale unbounded outdoor scenes while ensuring high-quality image rendering. 

\end{itemize}

\section{Related Works}
\textbf{Conventional Multi-Sensor Fusion SLAM}.
Conventional SLAM systems typically rely on filter-based~\cite{fastlio,fastlio2, fastlivo,zheng2024fast} or graph-based~\cite{zhang2017low,shan2020lio,shan2021lvi} solvers for ego state estimation, which generally allows for robust, accurate, and fast localization performance. In recent developments, LiDAR-Inertial-Visual SLAM systems have emerged as a versatile framework, which offers high-rate motion compensation, precise geometric structures, and rich texture information~\cite{fastlivo,r2live,r3live,shan2021lvi} that effectively mitigate issues in challenging environments.  Nevertheless, these methods still primarily concentrate on geometric mapping, which limits their capacity for achieving photometric image rendering.

\textbf{NeRF-based SLAM}.
NeRF uses a multi-layer perception (MLP) neural network and volume rendering to implicitly learn a 3D scene, with the network storing the structural information of the scene~\cite{Mildenhall20eccv_nerf}. However, this approach leads to a significant increase in computational time in large-scale environments.
Moreover, the primary drawback of the NeRF-based method is its extremely poor generalization capability across different scenes. Some methods integrate voxel~\cite{zhu2022nice,yang2022vox} or hash-tri-plane~\cite{wang2023co,johari2023eslam,sandstrom2023point} representations with MLPs as a way to construct maps, enabling reconstruction over larger areas. 
NeRF-SLAM~\cite{rosinolnerf2023} utilizes the dense depth maps estimated from DROID-SLAM~\cite{teed2021droid} as additional information to supervise the training of Instant-NGP~\cite{muller2022instant}. 
H$_2$-Mapping~\cite{jiang2023h} and H$_3$-Mapping~\cite{jiang2024h3} utilize hierarchical hybrid representations to enhance the reconstruction and achieve impressive performance in indoor scenes via RGB-D camera. NeRF-LOAM~\cite{Deng2023NeRFLOAMNI} and EINR~\cite{yan2023effi} gains better performance in large-scale environments, but they mainly focus on geometry structure reconstruction instead of photo-realistic rendering.
Despite these aforementioned advancements, implicit representation methods still face challenges regarding real-time performance, particularly in outdoor scenes.

\textbf{3DGS-based SLAM}. To improve computational efficiency,
3DGS utilizes sphere-based explicit scene representation, replacing time-consuming ray-based volume rendering with fast rasterization. This method accelerates NVS and produces promising rendering quality. Aimed at photorealistic mapping, methods like SplaTAM~\cite{keetha2024splatam}, Gaussian-SLAM~\cite{yugay2023gaussian}, and GS-SLAM~\cite{yan2024gs} showcase the substantial benefits of 3DGS over conventional map representations in SLAM tasks. MonoGS~\cite{matsuki2024gaussian} utilizes a single RGB sensor and introduces geometric regularization to address ambiguities in incremental reconstruction, demonstrating promising results with monocular images in small-scale scenes.
Additionally, Photo-SLAM~\cite{huang2024photo} leverages the classical visual odometry method ORB-SLAM3~\cite{campos2021orb} for precise state estimation and constructs a hybrid Gaussian map integrated with ORB features.
They typically rely on RGB-D/RGB camera sensors to achieve photo-realistic reconstruction in indoor scene.
On the other hand, DrivingGaussian~\cite{zhou2024drivinggaussian}, StreetGaussian~\cite{yan2024street}, and LIV-GaussMap~\cite{hong2024liv} realize photorealistic reconstructions based on multi-sensor fusion in ourdoor scenes. However, these approaches are all offline methods.
Gaussian-LIC~\cite{lang2024gaussian}, developed within a fusion SLAM framework~\cite{lang2023coco}, claims to provide high-quality photo-realistic mapping performance. 
However, it is reported that each keyframe iteration takes about one second, which is far slower than the 100 milliseconds requiring for real-time mapping.
Additionally, it struggles to process entire outdoor sequences due to GPU memory overflow, as it employs millions of 3D Gaussians to reconstruct the complete scene. MM3DGS SLAM~\cite{sun2024mm3dgs}, MM-Gaussian~\cite{wu2024mm}, and HGS-Mapping~\cite{wu2024hgs} complete multi-modal sensor fusion reconstruction in urban scenes. However, these methods are similar to offline method~\cite{hong2024liv}, which extend the optimization iteration to enhance performance metrics, yet it remains insufficient for achieving real-time capabilities.

\section{Methodology}
\label{sec:methodology}

\begin{figure}[tbp]
\centering
\includegraphics[width=\linewidth]{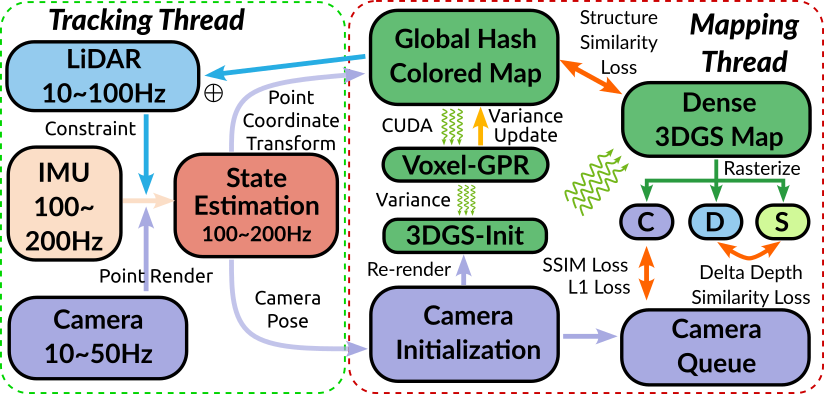}
\caption{The system processes input from point cloud data obtained from LiDAR, motion data from an IMU, and monocular color image captured by a camera. In the tracking thread, the ESIKF algorithm is employed to track motion, producing odometry output at the IMU frequency. In the mapping thread, the rendered color point cloud is utilized for Voxel-GPR, after which the initialized 3D Gaussian data is integrated into a dense 3D Gaussian map for rendering optimization. The final output is a high-quality, dense 3D Gaussian map. Notations C, D, and S denote the rasterized color image, depth image, and silhouette image, respectively. $\oplus$ represents the global hash-colored map, which provides neighboring query points for the recently scanned LiDAR points.}
\label{fig:overview}
\vspace{-1em}
\end{figure}

In this work,  we rely on the online LiDAR-Inertial-Visual fusion SLAM framework~\cite{r3live, yuan2024sr} for robust state estimation and point coordinate transformation. To address sparsity issue of LiDAR point cloud, we introduce voxel-level GPR (Voxel-GPR) in~\mysecref{sec:gp}. In~\mysecref{sec:sr_init}, we introduce a method to estimate the initial scale and rotation parameters of 3D Gaussians based on Voxel-GPR to speed up the convergence. Following this, we present our iterative system in~\mysecref{sec:system}, which encompasses updates to the GPR parameters, map expansion, similarity regularization, and implementation details. 
The brief overview of our framework is illustarted in~\myfigref{fig:overview}. 

\begin{algorithm}[htb]
\caption{Voxel-GPR for processing single-frame collected point clouds.}
\label{alg:voxelgpr}
\KwIn{$P_{f},\ \mathcal{C}\ (Global\ Stored\ Point\ Cloud\ in\ Voxels)$}
\KwOut{$P_{f*}$}
\DontPrintSemicolon
$\mathcal{S}_{updated}$ $\leftarrow$ \Call{Alg.STORE}{$P_{f}$,\enspace$\mathcal{C}$}\;
(CUDABatched) \For{$H_{p} $ $\mathbf{in}$ $\mathcal{S}_{updated}$}{
	${\mathcal{P}_{\alpha}}$ = $\mathcal{S}_{updated}[H_{p}]$ $\rightarrow$ $points$\;
	$\mathbf{V}_{\alpha} = \Call{Alg.PCA}{{\mathcal{P}_{\alpha}}} $\;
\Switch{$\mathbf{V}_{\alpha}$}{
       \uCase{$\mathbf{X}$}{
$\mathbf{f}_\alpha$ $\leftarrow$ $\mathcal{P}_{\alpha}^x$, $\mathbf{x}_\alpha$ $\leftarrow$  $\begin{bmatrix}\mathcal{P}_{\alpha}^y
  &\mathcal{P}_{\alpha}^z
\end{bmatrix}^\top $ \;
$\mathbf{x}_{\alpha *}$ $\leftarrow$ \Call{Alg.MeshGrid}{ $\mathcal{P}_{\alpha*}^y$ $, \mathcal{P}_{\alpha*}^z$}
}
       \uCase{$\mathbf{Y}$}{
$\mathbf{f}_\alpha$ $\leftarrow$  $\mathcal{P}_{\alpha}^y$, $\mathbf{x}_\alpha$ $\leftarrow$  $\begin{bmatrix}\mathcal{P}_{\alpha}^z
  &\mathcal{P}_{\alpha}^x
\end{bmatrix}^\top $   \;
$\mathbf{x}_{\alpha *}$ $\leftarrow$ \Call{Alg.MeshGrid}{ $\mathcal{P}_{\alpha*}^z$ $, \mathcal{P}_{\alpha*}^x$}
}
       \uCase{$\mathbf{Z}$}{
$\mathbf{f}_\alpha$ $\leftarrow$ $\mathcal{P}_{\alpha}^z$, $\mathbf{x}_\alpha$ $\leftarrow$ $\begin{bmatrix}\mathcal{P}_{\alpha}^x
  &\mathcal{P}_{\alpha}^y
\end{bmatrix}^\top $ \;
$\mathbf{x}_{\alpha *}$ $\leftarrow$  \Call{Alg.MeshGrid}{ $\mathcal{P}_{\alpha*}^x$ $, \mathcal{P}_{\alpha*}^y$}
}
}
$\mathbf{f}[H_p]$ $\leftarrow$ $\mathbf{f}_\alpha$, 
$\mathbf{x}[H_p]$ $\leftarrow$ $\mathbf{x}_\alpha$,
$\mathbf{x}_{*}[H_p]$ $\leftarrow$ $\mathbf{x}_{\alpha *}$
}

(CUDABatched) $\mathbf{K} \leftarrow \kappa(\mathbf{x},\mathbf{x}_*)$\;
(CUDABatched) $\mathbf{K_*} \leftarrow \kappa(\mathbf{x},\mathbf{x}_*)$\;
(CUDABatched) $\mathbf{K_{**}} \leftarrow \kappa(\mathbf{x_*},\mathbf{x}_*)$\;
(CUDABatched) $\mu_{*}, \boldsymbol{\Sigma}_{*} \leftarrow$ \Call{Alg.Solve}{} \myeqref{eq:calculate_f}\;
$P_{f*}$ = $(\mathbf{f}_*,\enspace \mathbf{x}_{*})$
\end{algorithm}

\subsection{Voxel-Level Gaussian Process Regression}
\label{sec:gp}

A series of studies~\cite{hong2024liv, lang2024gaussian, zheng2024fast} demonstrate that uneven point clouds in the 3DGS framework lead to memory inefficiency and reduced optimization. Offline methods can directly utilize original point clouds since occlusion and insufficient supervision are managed by density control algorithms. However, in dynamic scenarios, rapid perspective changes cause uneven gradient supervision, hindering quick convergence (refer to ablation experiments in~\mysecref{sec:abla}). To address this, we introduce Voxel-GPR, which employs GPR to uniformly transform point clouds at the voxel level, enhancing the efficiency of 3D Gaussian map optimization.


For each scanned voxel in continuous frame, we utilize Voxel-GPR for point cloud completion and to generate an evenly sampled point cloud $P_{f*}$ based on the uneven input $P_{f}$. The pseudocode for Voxel-GPR implemented using CUDA is provided in \myalgref{alg:voxelgpr}. 
We first apply a hash function to $P_{f}$ and simultaneously record the visited voxel hash \textit{points} as $\mathcal{S}_{updated}$.
The pseudocode from $P_{f}$ to $\mathcal{S}_{updated}$ in line 1 is introduced in the supplementary materials.
Then we perform Voxel-GPR processing on each voxel parallelly. 

For the $\alpha$th voxel, when the number of points contained is sufficient to meet the point count threshold $\tau$, we conduct Voxel-GPR operations on it. Let $n$ denote the number of points in $\mathcal{P}_{\alpha}$.
$\mathcal{P}_{\alpha}\in\mathbb{R}^{n\times 3}$ represents the \textbf{training} subset of point clouds contained within this voxel.
Inspired by~\cite{ruan2020gp}, we compute the eigenvector of point cloud $\mathcal{P}_{\alpha}$ by principal component analysis (PCA), and then determine the angles between $\mathbf{V}_\alpha$ and the three axes $\mathbf{X}$, $\mathbf{Y}$, and $\mathbf{Z}$, respectively. Then, the axis with the smallest angle is defined as the \textit{value axis} and the projection of $\mathcal{P}_{\alpha}$ onto the \textit{value axis} is denoted as $\mathbf{f}_{\alpha}$ ($\mathbf{f}_{\alpha}\in\mathbb{R}^{n}$), while the other two axes are considered as the \textit{parameter axes} and the projection of $\mathcal{P}_{\alpha}$ onto the \textit{parameters axis} is denoted as $\mathbf{x}_{\alpha}$ ($\mathbf{x}_{\alpha}\in\mathbb{R}^{n\times 2}$).
Note that $\mathbf{f}_\alpha=\mathbf{y}_\alpha+\epsilon$, where $\mathbf{y}_\alpha$ is noise-free observations, noise $\epsilon \sim  \mathcal{N}(0,\sigma^2)$ is independently added to each observation, where $\sigma$ is the pre-defined sensor variance (or the updated variance from \mysecref{sec:system}). $\mathcal{P}_{\alpha}$ in $\alpha$th voxel is assigned a random variable $\mathbf{f}_\alpha$ and the joint distribution is given by
\begin{equation}
p(\mathbf{f}_\alpha\mid\mathbf{x}_\alpha) \sim \mathcal{N}(\mathbf{f}_\alpha\mid\mu_\alpha,\mathbf{K}_\alpha)
\end{equation}
where $\mu_\alpha$ is the mean value in each dimension. $\mathbf
{K}_\alpha=\kappa(\mathbf{x}_\alpha,\mathbf{x}_\alpha)$ and $\kappa$ is a positive definite kernel function defined as $\kappa\left(\mathbf{x}_{\alpha i},\mathbf{x}_{\alpha j}\right)=\exp \left(-\lambda (\mathbf{x}_{\alpha i}-\mathbf{x}_{\alpha j})(\mathbf{x}_{\alpha i}-\mathbf{x}_{\alpha j})^\top\right)$,
which is a scale value representing the distance between variable $\mathbf{x}_{\alpha i}$ and $\mathbf{x}_{\alpha j}$, and $\lambda $ is a constant ($\lambda =1$ in our experiments).

\vspace{-1em}

\begin{figure}[htb]
    \centering
  \subfloat[\label{fig:voxelgpr}]{%
       \includegraphics[width=0.5\linewidth]{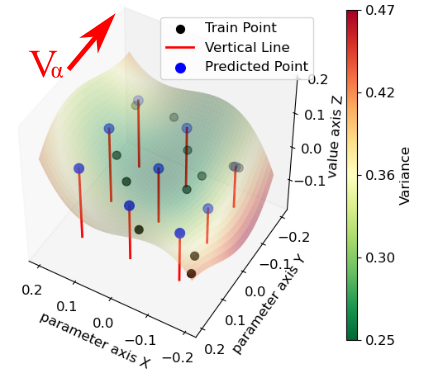}}
    \hfill
  \subfloat[\label{fig:sigmainit}]{%
       \includegraphics[width=0.5\linewidth]{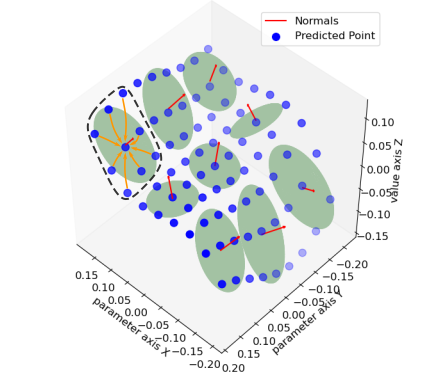}}
  \caption{(a) Illustration of Voxel-GPR for the $\alpha$th voxel involves processing voxels that contain a sufficient number $\tau$ of points. 
The depicted surface illustrates the distribution of the predicted points, where the variance decreases from areas marked in red to those in green. (b) Illustration depicts the initialization of 3D Gaussians. As indicated by the curved orange arrow, we consider the points within the neighborhood (black dash region). The green spheres represent the fitted 3D Gaussians.}
  \label{fig:type}
\vspace{-1em}
\end{figure}

Each side of $\alpha$th voxel consists of $n_s$ intervals. To generate the point cloud $\mathcal{P}_{\alpha*} \in \mathbb{R}^{n_{*} \times 3}$ ($n_{*}=n_s^2$), we first generate evenly spaced mesh grids $\mathbf{x}_\alpha$ on the plane determined by \textit{parameter axes}, which are then used as queries to the Voxel-GPR module to obtain $\mathbf{f}_{\alpha *} \in \mathbb{R}^{n_{*}}$ along \textit{value axis}.
Following the definition of a general GPR problem~\cite{rasmussen2003gaussian}, the joint distribution of observations $\mathbf{f}_\alpha$ and predictions $\mathbf{f}_{\alpha *}$ is again a Gaussian distribution which can be partitioned into
\begin{equation}
\label{eq:predict_f}
\binom{\mathbf{f}_\alpha}{\mathbf{f}_{\alpha *}} \sim \mathcal{N}\left(\mathbf{0},\left(\begin{array}{cc}
\mathbf{K_\alpha+\sigma^2\mathbf{I}} & \mathbf{K}_{\alpha *} \\
\mathbf{K}_{\alpha*}^\top & \mathbf{K}_{\alpha * *}
\end{array}\right)\right)
\end{equation}
where $\mathbf{K}_{\alpha*}=\kappa(\mathbf{x}_\alpha,\enspace\mathbf{x}_{\alpha *})$ and $\mathbf{K}_{\alpha**}=\kappa(\mathbf{x}_{\alpha *},\enspace\mathbf{x}_{\alpha *})$. With $n$ \textbf{training} data and $n_{*}$ \textbf{prediction} data, the predicted $\boldsymbol{\mu}_{\alpha *}$ and $\boldsymbol{\Sigma}_{\alpha *}$ are
\begin{align}
p\left(\mathbf{f}_{\alpha *} \mid \mathbf{x}_{\alpha *}, \mathbf{x}_\alpha, \mathbf{f}_\alpha\right) & = \mathcal{N}\left(\mathbf{f}_{\alpha *} \mid \boldsymbol{\mu}_{\alpha *},\enspace\boldsymbol{\Sigma}_{\alpha *}\right) \notag \\
\mathbf{f}_{\alpha *} &=\boldsymbol{\mu}_{\alpha *}  = \mathbf{K}_{\alpha *}^\top (\mathbf{K}_\alpha + \sigma^2\mathbf{I})^{-1} \mathbf{f}_\alpha \label{eq:calculate_f} \\
\boldsymbol{\Sigma}_{\alpha *} & = \mathbf{K}_{\alpha* *} - \mathbf{K}_{\alpha*}^\top (\mathbf{K}_\alpha + \sigma^2\mathbf{I})^{-1} \mathbf{K}_{\alpha*} \notag.
\end{align}

The prediction points $\mathcal{P}_{\alpha*}$ is given by $(\mathbf{f}_{\alpha *}$, $\mathbf{x}_{\alpha *})$, where $\mathbf{x}_{\alpha *}$ is the evenly spaced mesh grids. Moreover, $ \Sigma_{\alpha *}\in\mathbb{R}^{n_*\times n_*}$ is the estimated variance of $\mathcal{P}_{\alpha*}$. 
The prediction $\mathcal{P}_{\alpha*}$ and $\Sigma_{\alpha *}$ serve as representatives for $\alpha$th voxel and will be included in the initialization of the 3D Gaussians in~\mysecref{sec:sr_init}. Results of Voxel-GPR is depicted in~\myfigref{fig:voxelgpr}.

By leveraging CUDA's parallelization capabilities, we efficiently handle hundreds or thousands of voxels simultaneously, even in large-scale map expansions, it takes less than 30 milliseconds. 
The pseudocode for the CUDA batched algorithm ALG.SOLVE in line 21 of the Voxel-GPR algorithm is provided in the supplementary materials. 



\subsection{Efficient Initialization of 3D Gaussians}\label{sec:sr_init}
In our implementation of map management, each Gaussian $\mathcal{M}_k$ is defined by position $\mathbf{p}_k \in \mathbb{R}^{3}$, covariance matrix $\Phi_k \in \mathbb{R}^{3\times3}$, opacity $\Lambda_k \in \mathbb{R}$, and zero degree Spherical Harmonics (SHs) $Y_k \in \mathbb{R}^{3}$ per color channel. 
The scene of our dense 3DGS map $\mathcal{M}$ is 
\begin{equation}
\mathcal{M} = \{(\mathcal{M}_k:\mathbf{p}_k,\Phi_k,\Lambda_k, Y_k)\mid k=1,2,\cdots,N\}
\end{equation}
where $N$ is the number of 3D Gaussians in the dense map.

In the original implementation of 3DGS~\cite{kerbl20233d}, 3D covariance matrix $\Phi_k \in \mathbb{R}^{3\times3}$ of $k$th Gaussian is initiated as
\begin{equation}
\Phi_k = {R_k}{S_k}{S_k^{\top}}{R_k^{\top}}
\end{equation}
where $S_k$ is the scale vector calculated by finding the nearest distance to surrounding neighbors as described in~\cite{schoenberger2016sfm}, and $R_k$ is uniformly initialized as a constant and stored as a 4D quaternion. Benefited from Voxel-GPR, we develop an efficient initialization algorithm for calculating the covariance matrix $\Phi_k$ of $\mathcal{M}_k.$

In \mysecref{sec:gp}, the prediction side interval of the $\alpha$th voxel is $n_s$, and the prediction results within this voxel can be divided into $n_s \times n_s$ subgrids. In our actual code implementation, we predict the surrounding $n_r$ neighboring points for each interval additionally. Therefore, the number of predicted points on each side of the $\alpha$th voxel is $n_s \times n_r$. Next, we will discuss the algorithm to initialize 3D Gaussians for each subgrid. 

Points in the $\beta$th subgrid are defined as $\mathcal{G}^\beta = \{(\mathcal{P}_{f*}^\alpha)_i^\beta \in \mathbb{R}^{3}, \enspace w_i^\beta \in \mathbb{R}\}$, where $w_i^\beta$ is the weight coefficient for the $i$th point in the $\beta$th grid, given by $w_i^\beta = 1/\Sigma^\alpha_{*i}$, and $\Sigma^\alpha_{*i}$ is the covariance for the $i$th point. We use a sphere to regress $\mathcal{G}^\beta$ in this subgrid, as shown in \myfigref{fig:sigmainit}. Following this, the position $\mathbf{p}^\beta$ of the initial 3D Gaussian in this $\beta$th subgrid is given by
\begin{equation}
\mathbf{p}^\beta = \frac{\sum_{i=1}^{n_{r}^2} (\mathcal{P}_{f*}^\alpha)_i^\beta \cdot w_i^\beta}{\sum_{i=1}^{n_{r}^2} w_i^\beta}
\end{equation}
For scale and rotation parameters of each Gaussian, we fit neighbor by calculating the covariance matrix $\Phi^\beta$ in $\beta$th grid as
\begin{equation}
\Phi^\beta=\frac{\mathbf{Q}^\top \cdot diag(w_1^\beta, w_2^\beta, \dots, w_{n_r^2}^\beta) \cdot {\mathbf{Q}}}{\sum_{i = 1}^{n_{r}^2} w_i^\beta}
\end{equation}
where $\mathbf{Q}=(\mathcal{P}_{f*}^\alpha)^\beta - \mathbf{p}^\beta$.
The calculation of $Q$ described above involves subtracting $\mathbf{p}^\beta$ from each row of $(\mathcal{P}_{f*}^\alpha)^\beta$, resulting in $Q$ having a dimension of $n_r^2 \times 3$. The scale parameter $S^\beta$ in $\beta$th subgrid is given by
\begin{equation}
S^\beta = diag(\Phi^\beta)
\end{equation}
and $R^\beta$ of this 3D Gaussian is set to identity quaternion.
$S^\beta$ and $R^\beta$ are used to estimate initial shape of 3D Gaussians. For color information, we reproject $\mathbf{p}^\beta$ to current image by camera extrinsic, grab the color in this RGB image, and calculate the initial SHs $Y$.

By employing the aforementioned methods, we compute the parameters of $n_s \times n_s$ Gaussians for the $\alpha$th voxel. These 3D Gaussians serve as representatives of $\alpha$th voxel in space. Experimental results demonstrate that this approach significantly accelerates the optimization efficiency of 3D Gaussians while ensuring high-quality rendering.

\subsection{Iterative photo-realistic Mapping Framework}\label{sec:system}
\begin{figure}[!htb]
    \centering
  \subfloat[\label{fig:type:a}]{%
       \includegraphics[width=0.246\linewidth]{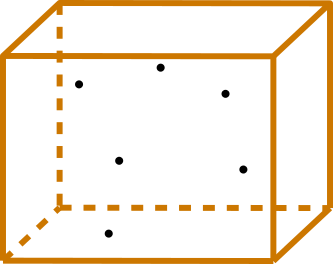}}
    \hfill
  \subfloat[\label{fig:type:b}]{%
       \includegraphics[width=0.246\linewidth]{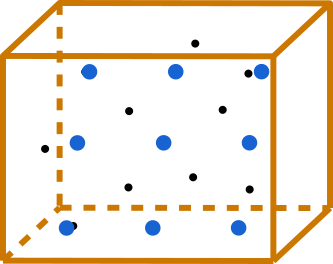}}
    \hfill
  \subfloat[\label{fig:type:c}]{%
       \includegraphics[width=0.246\linewidth]{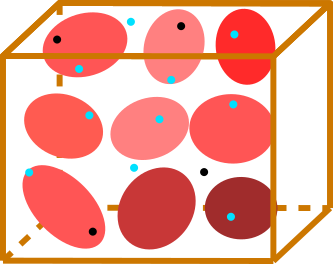}}
    \hfill
  \subfloat[\label{fig:type:d}]{%
       \includegraphics[width=0.246\linewidth]{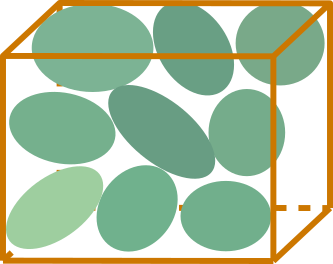}}
  \caption{
This illustration presents four types of voxels in hash voxel map. Small dots marked in black
represent new scanned points cloud from LiDAR, with their variance corresponding to the sensor's inherent noise. Larger dots highlighted in dark blue
are from the meshgrid, poised for processing via Voxel-GPR, while the small dots in light blue
have undergone variance updates. The ellipsoids are shaded in various colors, where each color signifies the magnitude of their variance.}
  \label{fig:type} 
\vspace{-1em}
\end{figure}

\textbf{Map Expansion and Covariance Update.}
For the voxels in the scene, we categorize them into four types: (a) unexplored voxels or those that do not meet the processing threshold $\tau $ (\myfigref{fig:type:a}); (b) voxels that meet the processing threshold but have not yet been added to the map (\myfigref{fig:type:b}); (c) voxels that are already included in the map but are still in an active state, meaning that their variance has not yet reached the convergence threshold $\eta$ and they requires further Voxel-GPR optimization (\myfigref{fig:type:c}); and (d) voxels that have completed convergence in Voxel-GPR module (\myfigref{fig:type:d}).

To increase the processing speed, we do not process types (a) and (d) by the Voxel-GPR module.
Our main focus is on processing the type (b) voxels, after which we expand these updated voxels to the dense map $\mathcal{M}$. The expansion is determined by whether the number of updated voxels exceeds a certain threshold, which can be found in the details of our code. 
For type (c) voxels, we update the original variance with the estimated value from the previous iteration. We then conduct the Voxel-GPR calculation again by stacking new observation points with the original points until the Voxel-GPR process converges for this voxel, at which point it transits to type (d).

\textbf{Rendering and Delta Depth Similarity Loss.}
Similar to~\cite{kerbl20233d}, we rasterize 3D Gaussians $\mathcal{M}$ to the observation camera with pose $\mathbf{T}$ and camera intrinsic $\pi$. The color of one pixel is rendered by $\mathcal{N}$ ordered 3D Gaussians in depth, photometric image $C$ is rendered as
\begin{equation}\label{equ:rendering}
C = \sum_{i\in \mathcal{N}}^{}c_i\alpha
_i\prod_{j = 1}^{i-1}(1-\alpha_j)  
\end{equation}
where $c_i$ represents color obtained by learned SHs coefficients and $\alpha_i$ is the density computed by multiplying 2D covariance $\sigma'$ with opacity $\Lambda_i$. 
We also rasterize per-pixel depth $D$ and silhouette image $S$ to determine that if a pixel contains information from the current map by
$\alpha$-blending rendering similar to~\cite{keetha2024splatam}.

In the delta depth similarity loss, we rasterize two depth image ($D_{\mathcal{F}_c}$, $D_{\mathcal{F}_{c+1}}$) and two silhouette image ($S_{\mathcal{F}_{c}}$, $S_{\mathcal{F}_{c+1}}$) to calculate delta depth similarity loss $\mathcal{L}_d$ between relative camera $\mathcal{F}_{c}=\{D_{\mathcal{F}_{c}}, S_{\mathcal{F}_{c}}, \pi_{\mathcal{F}_{c}}, \mathbf{T}_{\mathcal{F}_{c}}\}$ and $\mathcal{F}_{c+1}=\{D_{\mathcal{F}_{c+1}}, S_{\mathcal{F}_{c+1}}, \pi_{\mathcal{F}_{c+1}}, \mathbf{T}_{\mathcal{F}_{c+1}}\}$. $\mathcal{L}_d$ supervises the optimization of the 3DGS dense map through gradient back propagation. $\mathcal{L}_d$ is defined as
\begin{equation}
\label{equ:delta-depthloss}
\mathcal{L}_d = \|S_{\mathcal{F}_{c+1}}\circ D_{\mathcal{F}_{c+1}}- \pi_{\mathcal{F}_{c+1}} \mathbf{T}_{\mathcal{F}_{c+1}} \mathbf{T}^{-1}_{\mathcal{F}_c} \pi^{-1}_{\mathcal{F}_c} (S_{\mathcal{F}_c}\circ D_{\mathcal{F}_c}) \|_1  
\end{equation}
where $\circ$ denotes the operation applied element-wisely to the rendered depth image.

\textbf{Structure Similarity Loss.}\label{sec:simi}
For the scanned point cloud in the latest frame $P_f$, we introduce the similarity loss $\mathcal{L}_p$ to accelerate the optimization process. This loss measures the Euclidean distance between the current 3D Gaussian map $\mathcal{M} = \{\mathcal{M}_k:\mathbf{p}_k \in \mathbb{R}^{3} ,S_k \in \mathbb{R}^{3})\mid k=1,2,\cdots,m\}$ and the latest scanned point cloud frame $P_f$, $m$ and $n$ are the number of spheres and points, respectively. For each scanned point within the frame, we identify the nearest 3D Gaussian and calculate the Euclidean distance. Structure similarity loss $\mathcal{L}_p$ is defined as the mean of the nearest distances calculate by all points in $P_f$. $\mathcal{L}_p$ can be expressed as
\begin{equation}
\label{equ:simi}
\mathcal{L}_{p} = \frac{1}{n}\sum_{j=0}^{n} \min_{k\in\{1,2,\ldots,m\}} \left( \|(P_f)_j - \mathbf{p}_k\|_2 -  \bar{S_k} \right)
\end{equation}
where $\bar{S_k}$ is the mean value of $k$th 3D Gaussian scaling parameters.

\textbf{Training.}
The mapping thread in our system maintains a dense map composed of 3D Gaussians $\mathcal{M}$ and a continuous camera queue. We separate this camera queue into two parts: current visited camera window $\mathcal{Q}_{curr}$ and historical camera queue $\mathcal{Q}_{hist}$. $\mathcal{Q}_{curr}$ consists of $\mathcal{T}$ latested added cameras. We select $k_{curr}$ frames $\mathcal{F}_{curr}$ for map optimization during each iteration. To prevent catastrophic forgetting of historical data, each iteration also randomly selects $k_{hist}$ historical frame $\mathcal{F}_{hist}$ from $\mathcal{Q}_{hist}$ for joint optimization. 
Rasterized image loss, delta depth similarity loss, and structure similarity loss will be iterated to optimize the 3D Gaussian dense map by minimizing
\begin{equation}
\begin{array}{c}
\mathcal{L} = (1-\lambda_{s}) \|C - C_{gt}\|_1 + \lambda_{s}\mathcal{L}_{ssim} \\+  \lambda_{d}\sum_{\mathcal{F_*}\in\{\mathcal{F}_{curr},~\mathcal{F}_{hist}\}}^{} \mathcal{L}_d(\mathcal{F}_*,~\mathcal{F}_{*+1}) + \lambda_{p}\mathcal{L}_{p}
\end{array}
\end{equation}
where $C_{gt}$ and $C$ are the observed image and the image rendered by~\myeqref{equ:rendering}, respectively. $\mathcal{L}_{ssim}$ is an SSIM~\cite{wang2004image} term. $\mathcal{L}_d$ is the delta depth similarity loss calculated by $k_{curr}$ current frames $\mathcal{F}_{curr}$ and $k_{hist}$ historical frame $\mathcal{F}_{hist}$ in \myeqref{equ:delta-depthloss}. $\mathcal{L}_p$ is a structure similarity loss calculated in \myeqref{equ:simi}. $\lambda_{s}$, $\lambda_{d}$, and $\lambda_{p}$ are weights of SSIM loss, structure similarity loss, and delta depth similarity loss, respectively. 


\section{Experiments}

In this section, we first introduce the experimental setup in \mysecref{sec:expsetup}, including datasets, baseline, and parameter settings, etc. Then in \mysecref{sec:rendering}, we mainly compare the render performance of dense maps. \mysecref{sec:time-memory} provides the analysis of mapping time and GPU memory consumption of the framework. \mysecref{sec:abla} shows ablation experiments of the proposed framework.

\begin{table*}[htb]
\caption{Quantitative rendering performance comparison between radiance-field-based SLAM method~\cite{rosinolnerf2023}, 3DGS-based SLAM method~\cite{matsuki2024gaussian}, and original offline optimization~\cite{kerbl20233d} on R$^{3}$LIVE dataset~\cite{r3live}, FAST-LIVO dataset~\cite{fastlivo}, NTU dataset~\cite{nguyen2022ntu}, and Botanic Garden dataset~\cite{liu2023botanicgarden}. The results ranked from best to worst are highlighted as \colorbox[HTML]{c0e2ca}{first}, \colorbox[HTML]{fff5b3}{second}, and \colorbox[HTML]{ffd9b3}{third}.}
\label{tab:rendering}
\resizebox{\linewidth}{!}{
\begin{tabular}{cccccccccccccc}
\hline
                              &                                                                        & \multicolumn{3}{c}{Nerf-SLAM~\cite{rosinolnerf2023}}                                                                  & \multicolumn{3}{c}{MonoGS~\cite{matsuki2024gaussian}}                                                                     & \multicolumn{3}{c}{3DGS~\cite{kerbl20233d}}                                                                       & \multicolumn{3}{c}{Ours}                                                                       \\ \cline{3-14} 
\multirow{-2}{*}{Sequence}    & \multirow{-2}{*}{\begin{tabular}[c]{@{}c@{}}LiDAR\\ Type\end{tabular}} & PSNR↑                           & SSIM↑                          & LPIPS↓                         & PSNR↑                           & SSIM                         ↑ & LPIPS↓                         & PSNR↑                           & SSIM                         ↑ & LPIPS↓                         & PSNR↑                           & SSIM↑                          & LPIPS↓                         \\ \hline
\textit{hku\_campus\_seq\_00} & Livox Avia                                                             & \colorthird13.232 & \colorthird0.410 & 0.653                         & 12.142                         & 0.368                         & \colorthird0.608 & \colorsecond21.744 & \colorsecond0.719 & \colorsecond0.302 & \colorfirst22.430 & \colorfirst0.719 & \colorfirst0.247 \\
\textit{Visual\_Challenge}    & Livox Avia                                                             & 12.981                         & 0.408                         & 0.592                         & \colorthird13.478 & \colorthird0.579 & \colorthird0.414 & \colorsecond18.552 & \colorsecond0.545 & \colorsecond0.378 & \colorfirst21.806 & \colorfirst0.717 & \colorfirst0.289 \\
\textit{eee\_02}              & Ouster-16                                                              & 8.316                          & 0.320                         & 0.693                         & \colorthird11.632 & \colorthird0.407 & \colorthird0.514 & \colorsecond20.388 & \colorsecond0.686 & \colorsecond0.382 & \colorfirst20.718 & \colorfirst0.700 & \colorfirst0.319 \\
\textit{1005\_00}             & Livox Avia                                                             & 9.790                          & 0.362                         & 0.750                         & \colorthird14.563 & \colorthird0.602 & \colorthird0.406 & \colorfirst22.167 & \colorsecond0.657 & \colorsecond0.380 & \colorsecond21.123 & \colorfirst0.679 & \colorfirst0.258 \\\hline
\end{tabular}}
\vspace{-1em}
\end{table*}

\subsection{Experiment Setup}
\label{sec:expsetup}

\noindent\textbf{Datasets.}
We carry out experiments on four public LiDAR Inertial-Visual datasets, including R${^3}$LIVE~\cite{r3live} dataset, FAST-LIVO~\cite{fastlivo} dataset, NTU-VIRAL~\cite{nguyen2022ntu} dataset, and Botanic Garden~\cite{liu2023botanicgarden} dataset. The first two datasets are collected within the campuses of HKU and HKUST using a handheld device equipped with a Livox Avia LiDAR at 10\,Hz and its builtin IMU at 200\,Hz, and a 15\,Hz RGB camera (image resolution: 640$\times$512). The NTU-VIRAL~\cite{nguyen2022ntu} dataset is collect by Ouster-16 multi-line spinning LiDAR, which is more sparse than Livox LiDAR and the image resolution is 752$\times$480. The Botanic Garden~\cite{liu2023botanicgarden} dataset is collected by a wheeled robot traversing through a luxuriant botanic garden, 
point clouds from both multi-line spinning LiDAR Velodyne VLP-16 and Livox Avia LiDAR are collected and image resolution is 480$\times$300.

\noindent\textbf{Baselines and Metrics.}
We compare our method with the existing state-of-the-art NeRF-based dense visual SLAM NeRF-SLAM~\cite{rosinolnerf2023} and 3DGS-based SLAM~\cite{matsuki2024gaussian}. Additionally, the original offline implementation 3DGS~\cite{kerbl20233d} is also used as a baseline for comparison. It should be noted that RGB-D methods~\cite{yan2024gs, keetha2024splatam, yugay2023gaussian, huang2024photo, sucar2021imap} are not suitable in outdoor scenes due to the poor performance, so these methods are not included in comparison. We evaluate the rendering performance using PSNR, SSIM~\cite{wang2004image}, and LPIPS~\cite{zhang2018unreasonable}. Note that we run all the baselines on datasets and calculate the mean metric among all observation images. 


\noindent\textbf{Implementation Details.} Our framework is implemented by CUDA/C++ using the LibTorch framework~\cite{libtorch} under Robot Operating System (ROS)~\cite{quigley2009ros}, incorporating CUDA code for Gaussian Splatting and trained on a desktop PC with a 5.50\,GHz Intel Core i9-13900HX CPU, 64\,GB RAM, and a NVIDIA RTX 4060 Laptop 8\,GB GPU. In all scequences, the hyprparameters used in our experiments are listed in the supplementary materials.

\begin{figure}[!htb]
\centering
\includegraphics[width=\linewidth]{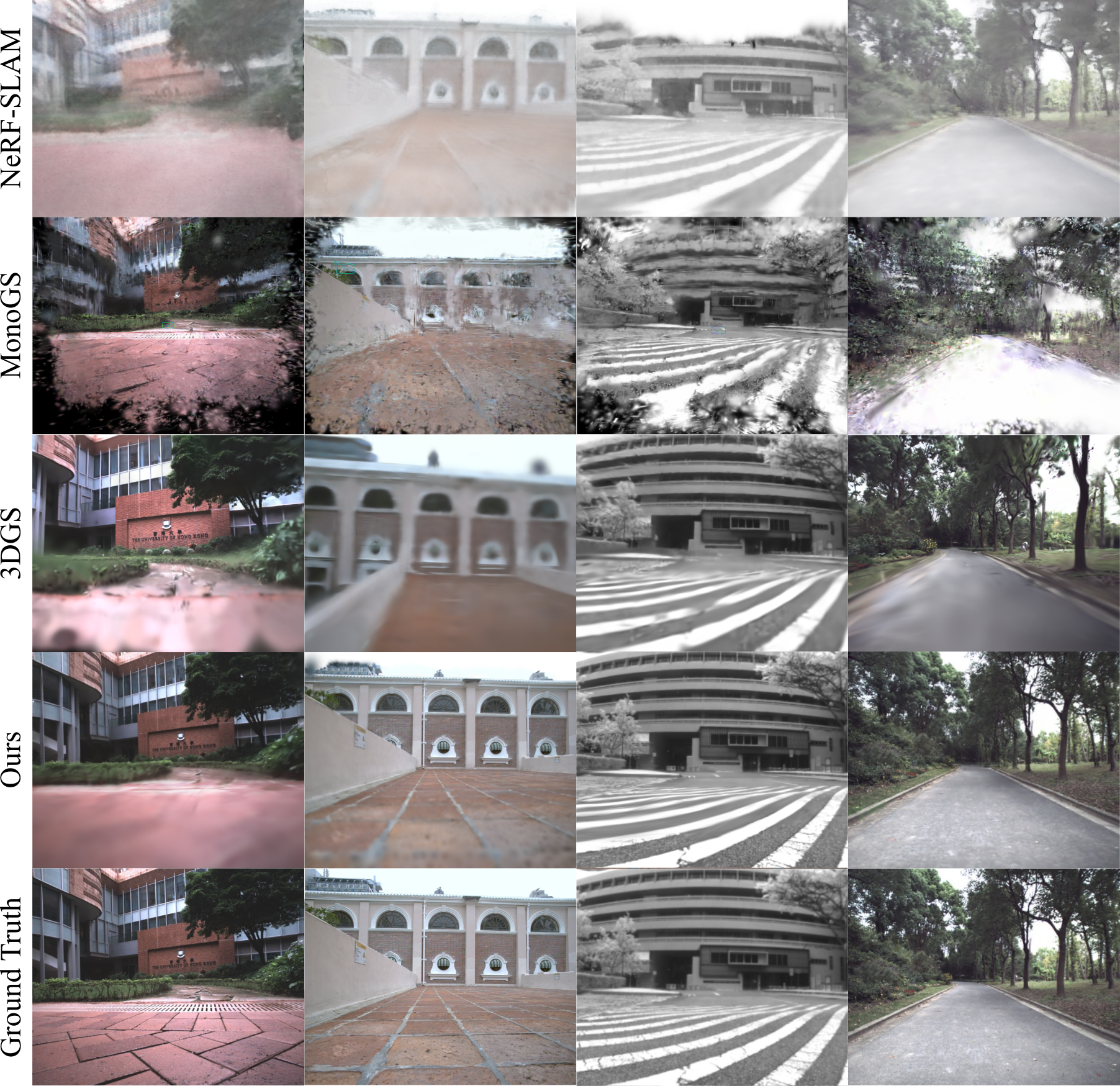}
\caption{Illustration of the rendering performance on the NeRF-based method~\cite{rosinolnerf2023} and the 3DGS-based method~\cite{matsuki2024gaussian,kerbl20233d} on sequence \textit{hkust\_campus\_seq\_00}, \textit{Visual\_Challenge}, \textit{eee\_03},
\textit{1005\_00}, respectively.
}
\label{fig:render}
\vspace{-1em}
\end{figure}

\subsection{Rendering Evalution}
\label{sec:rendering}
We conduct rendering comparisons using the 3DGS original implementation~\cite{kerbl20233d}, NeRF-SLAM~\cite{rosinolnerf2023}, and Gaussian Splatting SLAM~\cite{matsuki2024gaussian}. All of these methods rely solely on images as supervision. Due to limitations in GPU memory and poor tracking performance, these methods~\cite{kerbl20233d, rosinolnerf2023, matsuki2024gaussian} are unable to complete the reconstruction task of the entire environment. To facilitate comparison, we reduce the size of the reconstruction scene and use camera poses and depth maps calculated by COLMAP~\cite{schoenberger2016sfm} as input for methods~\cite{kerbl20233d, rosinolnerf2023, matsuki2024gaussian}. The camera pose in our framework is estimated by multi-sensor fusion odometry. The reconstruction sequence in the scene is limited to within 100 frames, and the reconstruction time is limited to within 5 minutes. The comparision of rendering performance are shown in \mytabref{tab:rendering} and \myfigref{fig:render}. It can be observed that due to the strict time constraints, the reconstruction results of NeRF-SLAM~\cite{rosinolnerf2023} and MonoGS~\cite{matsuki2024gaussian} are relatively poor. Although 3DGS performs well in some scenes, its offline optimization performs poorly on NVS. More details of rendering experiments can be found in the supplementary materials.

\begin{figure}[htb]
\centering
\includegraphics[width=\linewidth]{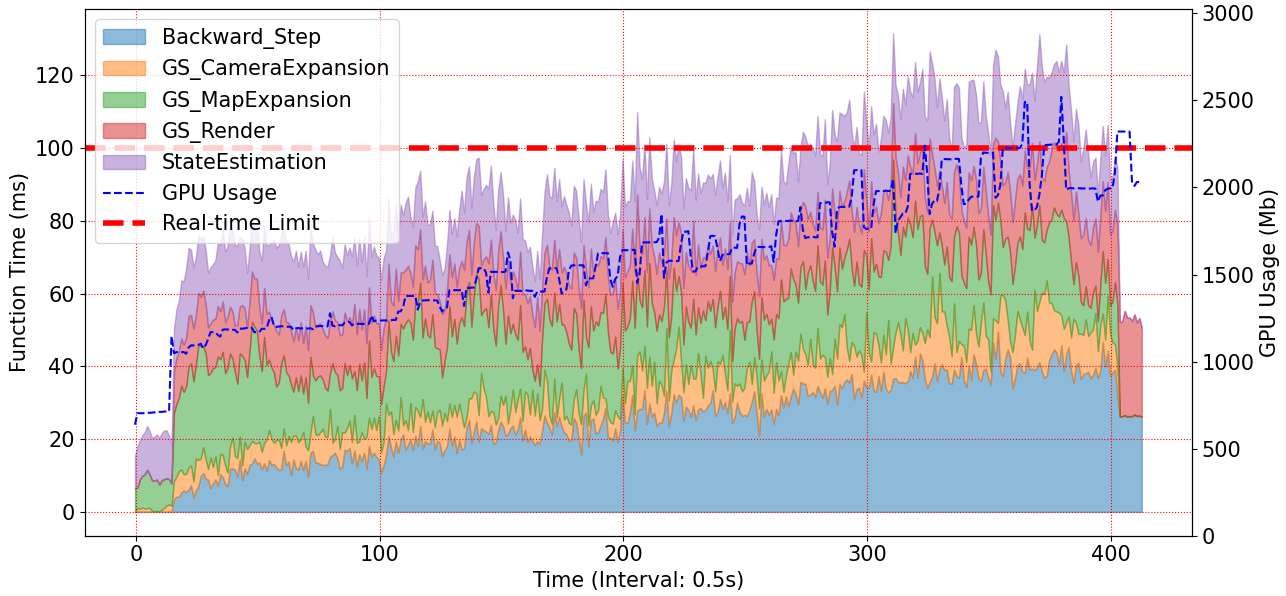}
\caption{Time consumption and GPU memory consumption on Botanic Garden sequence \textit{1018\_13}.}
\label{fig:timeanalyse}
\vspace{-1em}
\end{figure}

\begin{table}[!htb]
\caption{Duration (DT), mapping Time (MT), count of 3D Gaussians, and maximum memory cost (Mem) in photo-realistic mapping in different sequences.}
\centering
\label{tab:dura-gs-gpu}
\resizebox{0.85\linewidth}{!}{
\begin{tabular}{@{}ccccc@{}}
\toprule
                   Sequence           & DT (s) & MT (s) & Count     & Mem (Mb) \\ \midrule
\textit{hku\_campus\_seq\_00} & 202   & 202       & 1,209,666 & 2495     \\
\textit{Visual\_Challenge}    & 162    & 162      & 353,334   & 1353     \\
\textit{eee\_02}              & 321    & 321      & 758,213   & 1523     \\
\textit{1005\_00}             & 611    & 612        & 2,177,464 & 3492     \\ \bottomrule
\end{tabular}
}
\vspace{-1em}
\end{table}

\subsection{Runtime and Memory Cost Analysis}
\label{sec:time-memory}

The real-time performance criterion in our experiments is that all collected image frames are processed, meaning that the processing time for each frame $t$ should satisfy $t<=\mathcal{D}/\mathcal{C}$, where $\mathcal{D}$ is the duration of collected dataset and $\mathcal{C}$ is the size of added tracking camera queue.
The fill-between plot in mapping time consumption over time in the Botanic Garden sequence \textit{1018\_13 (duration: 208\,s)} is shown in \myfigref{fig:timeanalyse}. It can be observed that as time progresses, the overall system consumption time remains mostly below the real-time limit. The time consumption graph on longer sequences can be found in the supplementary materials. 

As the mapping process advances, the consumption of GPU memory escalates with an increasing number of 3D Gaussians. All sequences can be fully processed through photo-realistic mapping on an 8\,GB GPU when $n_s=3$. \mytabref{tab:dura-gs-gpu} outlines experiment results on various sequences. 


\subsection{Ablation Study}
\label{sec:abla}
\begin{figure}[!htb]
    \centering
  \subfloat[w/o Voxel-GPR\label{fig:abla:no-gp}]{%
       \includegraphics[width=0.33\linewidth]{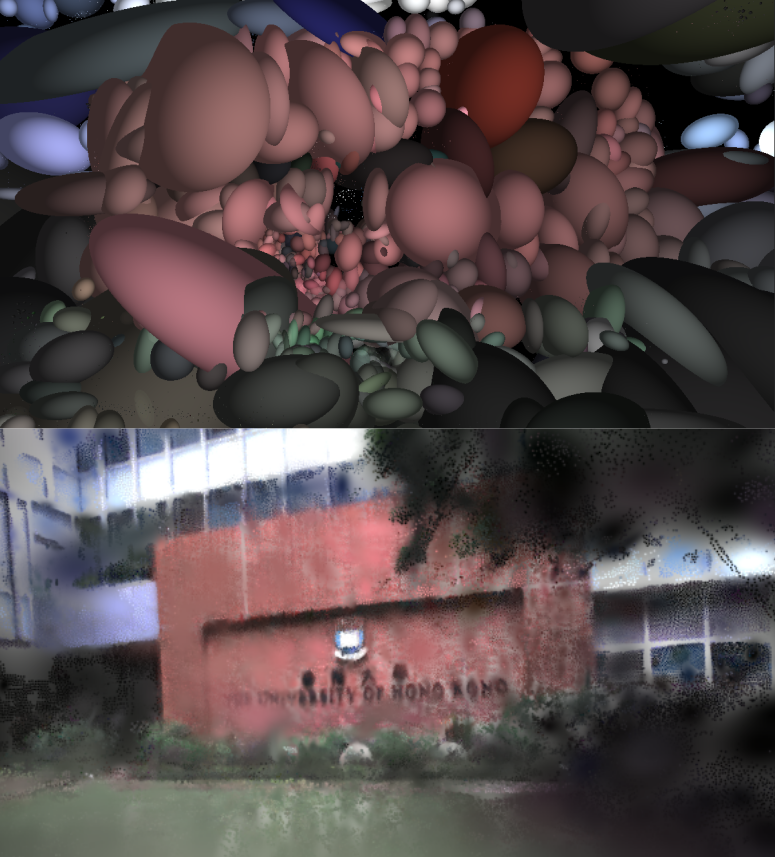}}
    \hfill
  \subfloat[w/o our initialization\label{fig:abla:no-covinit}]{%
       \includegraphics[width=0.33\linewidth]{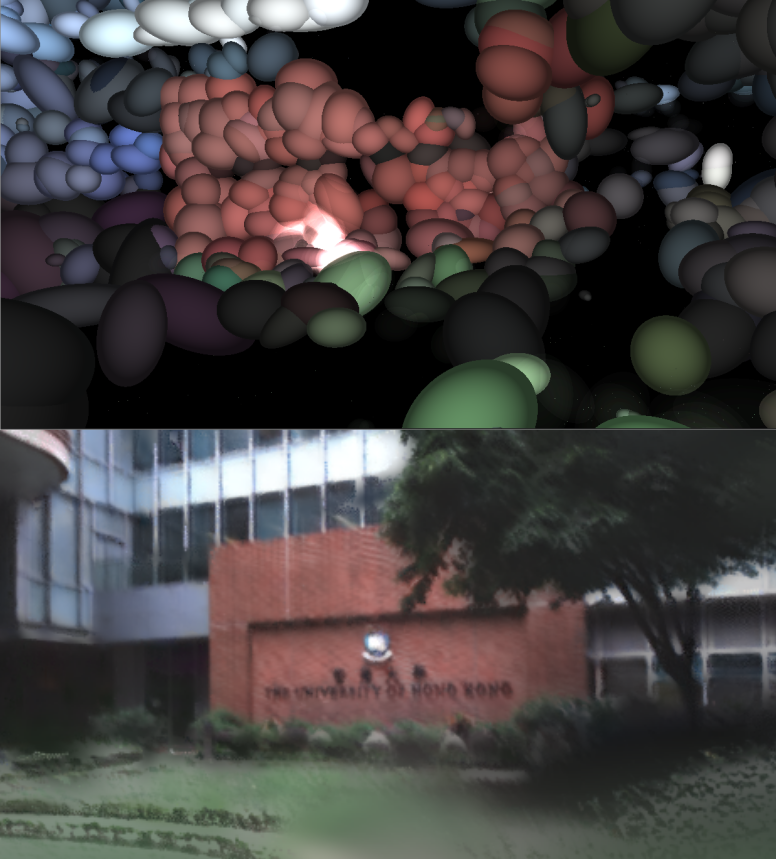}}
    \hfill
  \subfloat[w/o SS\label{fig:abla:no-simi}]{%
       \includegraphics[width=0.33\linewidth]{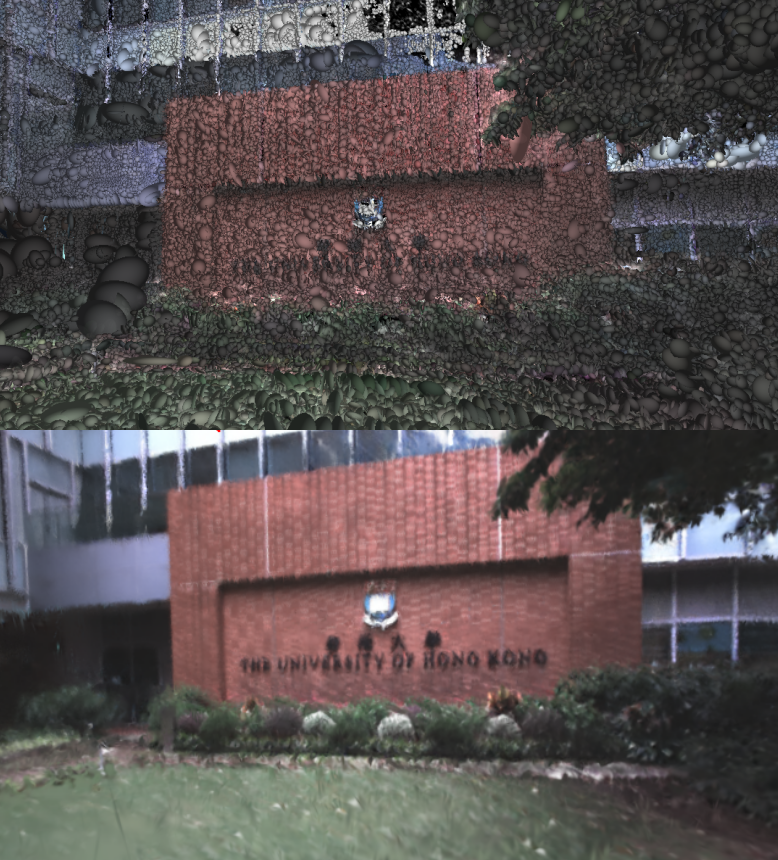}}
       \\
  \subfloat[w/o DDS\label{fig:abla:no-dd}]{%
       \includegraphics[width=0.33\linewidth]{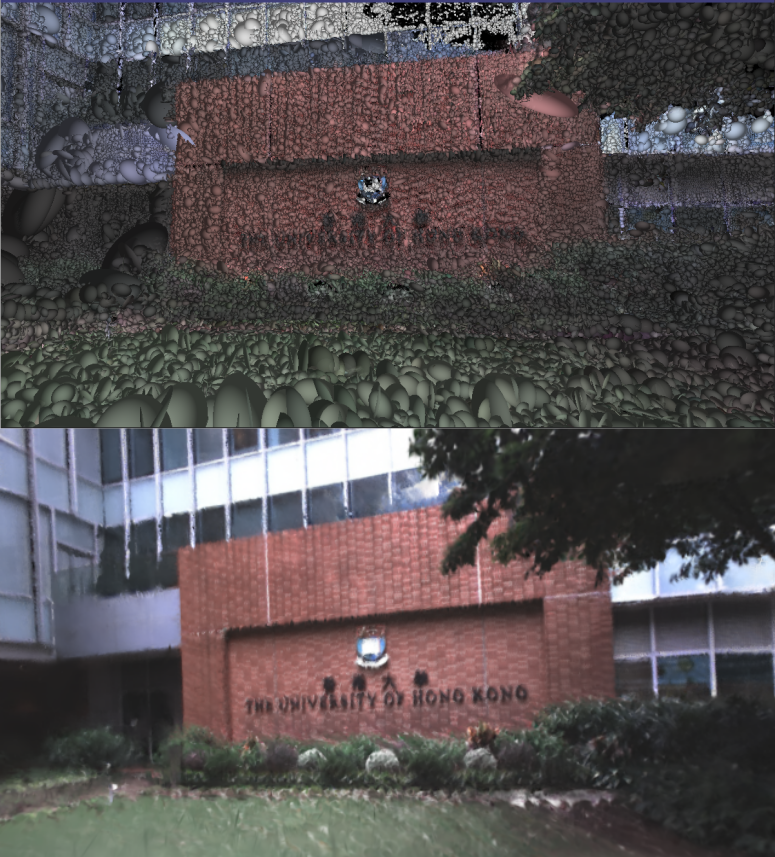}}
   	\hfill
  \subfloat[Full ($n_s=4$)\label{fig:abla:n3}]{%
       \includegraphics[width=0.33\linewidth]{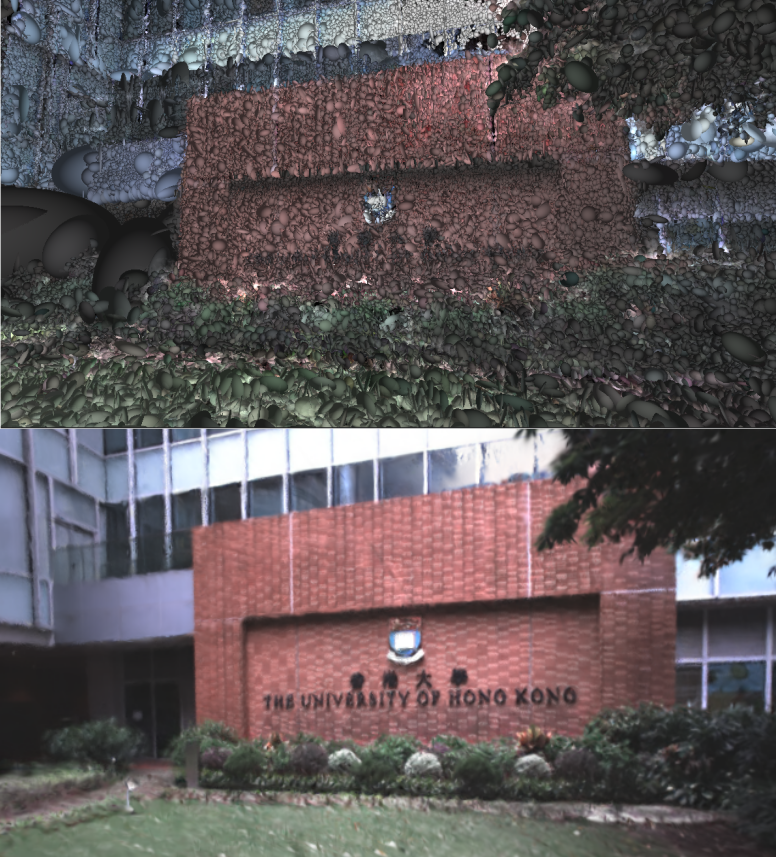}}
    \hfill
  \subfloat[Full ($n_s=3$)\label{fig:abla:n4}]{%
       \includegraphics[width=0.33\linewidth]{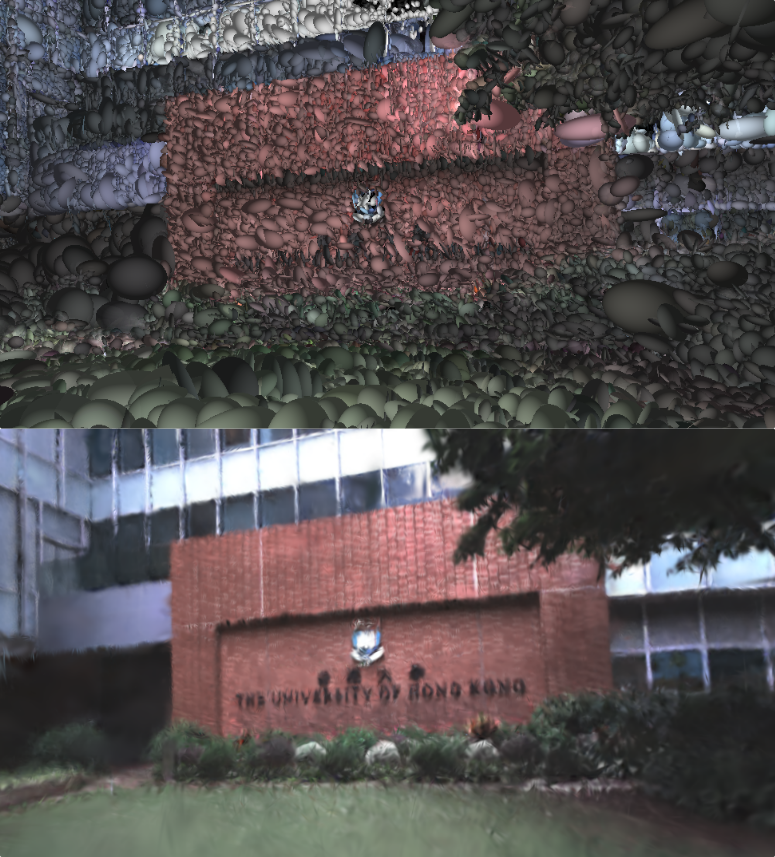}}
    \hfill
  \caption{(a) Similar to~\cite{hong2024liv, lang2024gaussian}, each collected point corresponds to a generated 3D Gaussian.
(b) The use of a uniform initialization method.
(c) There is no structural similarity loss (SS) for the single frame point cloud.
(d) There is no delta depth similarity loss (DDS) for the relative frame.
(e) The combined results of all the modules and losses are presented. Note that, $n_s=4$ in (a-e). 
(f) The combined results of all the modules ($n_s=3$).}
  \label{fig:abla}
\vspace{-1em}
\end{figure}

We conduct an ablation study on the Voxel-GPR module and the 3D Gaussians initialization algorithm module in our system, as well as the structure similarity loss and the delta depth similarity loss, as shown in \myfigref{fig:abla}. It can be seen that the construction of 3D Gaussians based on the original point cloud (\myfigref{fig:abla:no-gp}) does not consider the geometric dependencies in environment, which can lead to uneven distributions of 3D Gaussians. 
Moreover, achieving the desired optimization effect with \myfigref{fig:abla:no-covinit} and \myfigref{fig:abla:no-simi} takes 1.3 times longer, indicating that a reasonable initialization method can accelerate the convergence. 
By adding the structure similarity loss in \myfigref{fig:abla:no-dd} and the delta depth similarity loss in \myfigref{fig:abla:n3} based on \myfigref{fig:abla:no-simi}, the representation of details in the environment is enhanced, resulting in images that are closer to real collected images.

Quantitative ablation experiments are shown in \mytabref{tab:abla}, mainly comparing the differences in mapping time and mean rendering metrics on all observation images of different hyper parameters. The first and second rows of each sequence mainly compare the influence of parameter $\eta$ on mapping time and rendering metrics. Setting an appropriate value for $\eta$ can significantly impact mapping time for different LiDAR configurations. A smaller $\eta$ value results in longer mapping times, but does not provide a clear improvement in render metrics. 
The 2nd to 4th rows compare the effects of structure similarity loss and delta depth loss on render metrics. It can be observed that adding these two losses does not impact mapping time but can enhance render metrics. The 4th and 5th rows in \mytabref{tab:abla} and~\myfigref{fig:abla:n3}$\text{-}$~\myfigref{fig:abla:n4} compare the impact of $n_s$ on the results. In smaller scenes (e.g., \textit{1018\_13} in Botanic Garden~\cite{liu2023botanicgarden}, \textit{hkust\_campus\_seq\_00} in R$^3$LIVE~\cite{r3live}), $n_s$ does not affect mapping time and significantly improves render metrics, but in larger scenes (e.g., \textit{hku\_main\_building} in R$^3$LIVE~\cite{r3live}), $n_s$ may consume excessive time and GPU memory. This illustrates the trade-off between accuracy and speed. Expanded ablation experiments on more sequences can be refered in the supplementary materials.

\begin{table}[!htb]
\caption{Quantitative ablation comparison of hyperparameters in various sequences, such as duration (DT), mapping time (MT), interval of 3D Gaussians in voxel side $n_s$, convergence variance of Voxel-GPR $\eta$, structure similarity loss (SS), and delta depth similarity loss (DDS). 
}
\centering
\label{tab:abla}
\resizebox{0.9\linewidth}{!}{
\begin{tabular}{@{}ccccccccc@{}}
\toprule
                                                                                             Sequence & DT (s)   &MT (s)                 & $n_s$ & $\eta$ & SS & DDS &  PSNR  & SSIM                          \\ \midrule
                                                                                              &           & 1240            & 3    & 0.1  & \ding{55}    & \ding{55}   & 18.494                         & 0.637                         \\
                                                                                              &           & 202            & 3    & 0.3  & \ding{55}    & \ding{55}    & 19.005                         & \cellcolor[HTML]{FFD9B3}0.647 \\
                                                                                              &           & 202            & 3    & 0.3  & \ding{51}    & \ding{55}    & \cellcolor[HTML]{FFF5B3}19.944 & \cellcolor[HTML]{FFF5B3}0.648 \\
                                                                                              &           & 202            & 3    & 0.3  & \ding{51}    & \ding{51}    & \cellcolor[HTML]{FFD9B3}19.292 & 0.640                         \\
\multirow{-5}{*}{\textit{\begin{tabular}[c]{@{}c@{}}hku\_\\ campus\\ \_seq\_00\end{tabular}}} & \multirow{-5}{*}{202} & 202 & 4    & 0.3  & \ding{51}    & \ding{51}    & \cellcolor[HTML]{C0E2CA}20.143 & \cellcolor[HTML]{C0E2CA}0.672 \\ \midrule
                                                                                              &                       &891  & 3    & 0.2  & \ding{55}    & \ding{55}  &  14.754                         & 0.449                         \\
                                                                                              &                       & 208  &3    & 0.3  & \ding{55}    & \ding{55}  &  \cellcolor[HTML]{FFF5B3}16.039 & \cellcolor[HTML]{FFD9B3}0.475 \\
                                                                                              &                       & 208  &3    & 0.3  & \ding{51}    & \ding{55}  &  \cellcolor[HTML]{FFD9B3}15.988 & \cellcolor[HTML]{FFF5B3}0.488 \\
                                                                                              &                       & 208  & 3    & 0.3  & \ding{51}    & \ding{51}  & 15.732                         & 0.461                         \\
\multirow{-5}{*}{\textit{1018\_13}}                                                           & \multirow{-5}{*}{208} &208  & 4    & 0.3  & \ding{51}    & \ding{51}  &  \cellcolor[HTML]{C0E2CA}17.982 & \cellcolor[HTML]{C0E2CA}0.493 \\ \bottomrule
\end{tabular}
}
\vspace{-1em}
\end{table}

\section{Conclusion}
 In conclusion, this paper presents \textbf{GS-LIVM}, a novel real-time photo-realistic LiDAR-Inertial-Visual mapping framework with Gaussian Splatting. Our approach enables real-time photo-realistic mapping while ensuring high-quality image rendering in large, uncontrolled outdoor scenes. 
 In this work, we leverage Voxle-GPR to tackle the sparsity issue in LiDAR point clouds. The voxel-based 3D Gaussians map representation facilitates real-time photo-realistic mapping in large-scale unbounded outdoor scenes with the acceleration provided by custom CUDA kernels. Additionally, the framework is structured in a covariance-centered manner, where the estimated covariance is utilized to initialize the scale and rotation of 3D Gaussian, as well as update the parameters of the Voxle-GPR.
Experimental evaluations on various outdoor datasets validate that our algorithm achieves state-of-the-art performance in terms of mapping efficiency and rendering quality. Moreover, the source code is publicly available, facilitating further research and development in this area.

{
    \small
    \bibliographystyle{ieeenat_fullname}
    \bibliography{main}
}

\clearpage
\setcounter{page}{1}
\maketitlesupplementary

\section{Pseudocode for Preprocessing Algorithm ALG.STORE \& CUDA Batched Algorithm ALG.SOLVE in Voxel-GPR}
Preprocessing algorithm for latest scanned point cloud $P_{f}$ is shown in \myalgref{alg:supply:preprocess}.  Each point in $P_{f}$ is stored into a voxel structure, and the visited voxel information in this frame is recoeded simultaneously. Utilizing the cuBLAS scientific computing library within the GPU, the pseudocode for batch-solving \myeqref{eq:calculate_f} for all voxels is shown as \myalgref{alg:supply:solvef}.

\begin{algorithm}[htb]
\caption{Preprocessing for single-frame collected point clouds}
\label{alg:supply:preprocess}
\KwIn{$P_{f},\ \mathcal{C}\ (Global\ Stored\ Point\ Cloud\ in\ Voxels)$}
\KwOut{$\mathcal{S}_{updated}(Store\ Visited\ Voxels\ in\ This\ Frame)$}
\DontPrintSemicolon
$\mathcal{S}_{updated} \leftarrow [\ ]$\;

\For{$p $ $\mathbf{in}$ $P_{f}$}{
	$H_{p}$ = \Call{Alg.HASH}{$p^x$, $p^y$, $p^z$}\;
	\textbf{push} $p$ \textbf{to} $\mathcal{C}[H_{p}]$ $\rightarrow$ $points$\;
	\textbf{push} $H_{p}$ \textbf{to} $\mathcal{S}_{updated}$\;
	}
\end{algorithm}

\begin{algorithm}[htb]
\caption{Batch solving $\mu_*$ and $\Sigma_*$ in Voxel-GPR based on cuBLAS}
\label{alg:supply:solvef}
\KwIn{$\mathbf{f},~\mathbf{K},~\mathbf{K}_*,~\mathbf{K}_{**}$}
\KwOut{$\mu_*$,~$\Sigma_*$}
\DontPrintSemicolon
$\#~Add~Noise$\;
$~~~~~\mathbf{ky}$ $\leftarrow$ $\mathbf{K} + \sigma^2\mathbf{I}$\;

$\#~In\-place~LU~Decomposition$\;
$~~~~~$cublasSgetrfBatched(ky)\;

$\#~Calculate~Inverse~Matrix~of~\mathbf{ky}$\;
$~~~~~$kyInv $\leftarrow$ cublasSgetriBatched(ky)\;

$\#~Matrix~Multiplication$\;
$~~~~~k_{k*}$ $\leftarrow$ cublasSgemmBatched($\mathbf{K}_*$, kyInv)\;
$~~~~~\mathbf{f_*}$ $\leftarrow$ $\mu_*$ $\leftarrow$ cublasSgemmBatched($k_{k*}$, $\mathbf{f}$)\;
$~~~~~\Sigma_*$ $\leftarrow$ $\mathbf{K_{**}}$ - cublasSgemmBatched($k_{k*}$, $\mathbf{K}_*)$\;
\end{algorithm}

\section{Supplementary Experiments}
\subsection{Comparison with Fully Optimized Offline 3DGS}
\label{sec:supply:comparisonwith3dgs}
\begin{figure}[!htb]
    \centering
  \subfloat[3DGS~\cite{kerbl20233d}]{%
       \includegraphics[width=0.48\linewidth]{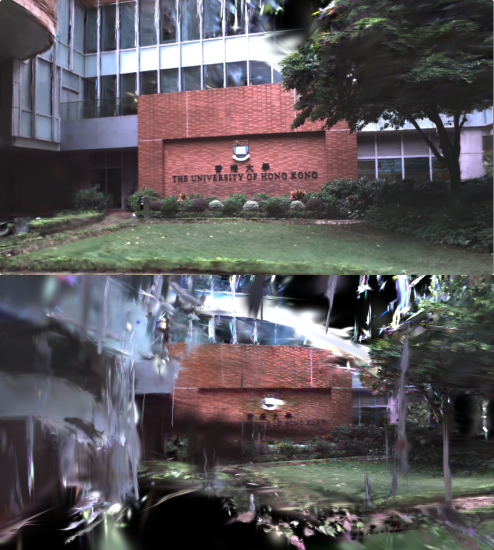}}
    \hfill
  \subfloat[Ours]{%
       \includegraphics[width=0.48\linewidth]{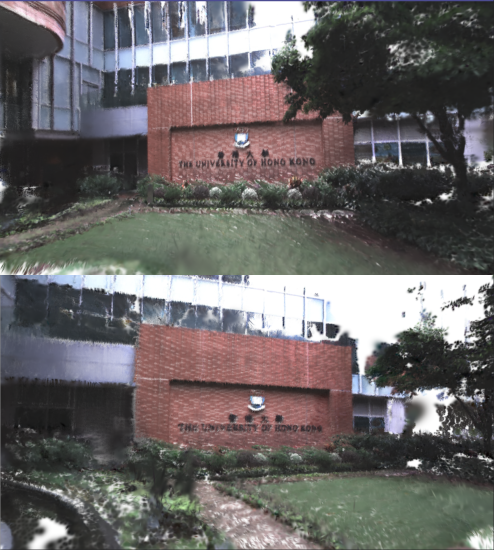}}
     \\
  \subfloat[3DGS~\cite{kerbl20233d}]{%
       \includegraphics[width=0.48\linewidth]{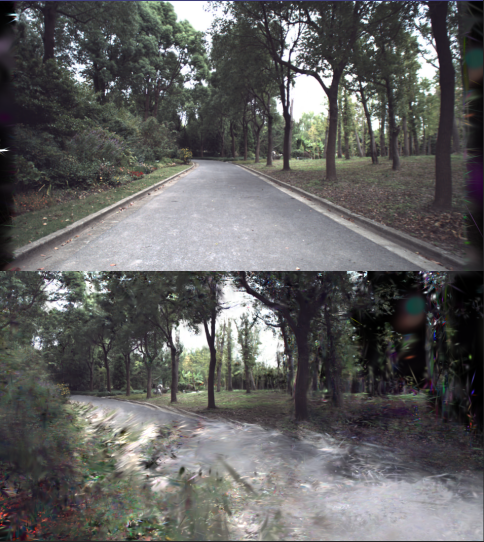}}
    \hfill
  \subfloat[Ours]{%
       \includegraphics[width=0.48\linewidth]{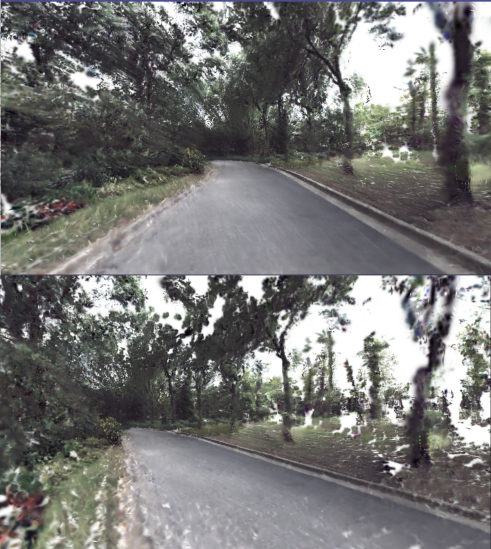}}
  \caption{In all subfigures, the top images represent a supervised perspective with image information, while the bottom images depict an unsupervised perspective with synthesized new viewpoints. Results in (a)(c) are obtained using the offline 3DGS method, while results in (b)(d) are obtained using our method.}
  \label{fig:supply:3dgs}
\end{figure}

In the rendering comparison presented in \myfigref{fig:render}, to ensure a fair comparison, we limit the training of the original 3DGS implementation to 3000 iterations. Interestingly, we discover that with sufficient optimization time, 3DGS indeed achieves better metrics than our method (\mytabref{tab:supply3dgs}). This is because 3DGS involves offline optimization over an extended period to overfit the training data. 
Nonetheless, limitations are encountered with NVS, as demonstrated in \myfigref{fig:supply:3dgs}.
Overfitting to the training dataset causes 3D Gaussians to extend beyond the observation point, which is the primary issue with offline methods. It can be observed that our method can overcome the drawbacks of overfitting.

\begin{table}[!htb]
\centering
\caption{Comparison with the fully optimized offline 3DGS~\cite{kerbl20233d}.}
\label{tab:supply3dgs}
\resizebox{\linewidth}{!}{
\begin{tabular}{p{1.5cm}cccccc}
\toprule
\multirow{2}{*}{Sequence} & \multicolumn{3}{c}{3DGS} & \multicolumn{3}{c}{Ours} \\
                          & PSNR↑     & SSIM↑    & LPIPS↓    & PSNR↑   & SSIM↑   & LPIPS↓  \\\midrule
\textit{hku\_seq\_00}                  &     27.827     &  0.861       &   0.204       &     22.430   & 0.719       &  0.247      \\
\textit{1005\_00}                  &     26.196     &  0.812       &   0.233       &     21.123   & 0.679       &  0.258      \\
\bottomrule
\end{tabular}
}
\end{table}

\begin{table*}[htb]
\caption{Quantitative rendering performance comparison on more datasets. The results ranked from best to worst are highlighted as \colorbox[HTML]{c0e2ca}{first}, \colorbox[HTML]{fff5b3}{second}, and \colorbox[HTML]{ffd9b3}{third}.}
\label{tab:supply:rendering}
\resizebox{\linewidth}{!}{
\begin{tabular}{cccccccccccccc}
\hline
                              &                                                                        & \multicolumn{3}{c}{Nerf-SLAM~\cite{rosinolnerf2023}}                                                                  & \multicolumn{3}{c}{MonoGS~\cite{matsuki2024gaussian}}                                                                     & \multicolumn{3}{c}{3DGS~\cite{kerbl20233d}}                                                                       & \multicolumn{3}{c}{Ours}                                                                       \\ \cline{3-14} 
\multirow{-2}{*}{Sequence}    & \multirow{-2}{*}{\begin{tabular}[c]{@{}c@{}}LiDAR\\ Type\end{tabular}} & PSNR↑                           & SSIM↑                          & LPIPS↓                         & PSNR↑                           & SSIM                         ↑ & LPIPS↓                         & PSNR↑                           & SSIM                         ↑ & LPIPS↓                         & PSNR↑                           & SSIM↑                          & LPIPS↓                         \\ \hline
 \textit{hkust\_campus\_02}    & Livox Avia                                                             & 10.728                         & 0.389                         & 0.716                         & \colorthird12.408 & \colorthird0.426 & \colorthird0.582 & \colorfirst17.629 & \colorfirst0.692 & \colorsecond0.412 & \colorsecond15.745 & \colorsecond0.634 & \colorfirst0.382 \\
 \textit{hku1}                 & Livox Avia                                                             & \colorthird9.232  & \colorthird0.384 & \colorthird0.716 & 8.928                          & 0.286                         & 0.750                         & \colorfirst19.548 & \colorfirst0.708 & \colorfirst0.328 & \colorsecond14.990 & \colorsecond0.600 & \colorsecond0.424 \\
 \textit{1018\_00}             & Velodyne VLP-16                                                         & 12.186                         & 0.387                         & 0.699                         & \colorthird12.721 & \colorthird0.634 & \colorthird0.417 & \colorfirst23.452 & \colorfirst0.720 & \colorfirst0.253 & \colorsecond16.971 & \colorsecond0.688 & \colorsecond0.338 \\
 \textit{private-360}          & Livox MID-360                                                          & \colorthird13.681 & \colorthird0.407 & \colorthird0.598 & 8.912                          & 0.263                         & 0.778                         & \colorfirst22.428 & \colorfirst0.693 & \colorfirst0.302 & \colorsecond17.777 & \colorsecond0.618 & \colorsecond0.439 \\ 
 \textit{private-pandar}          & Pandar XT-32 & \colorthird9.492 & \colorthird0.328 & \colorthird0.610 & 7.414                          & 0.289                         & 0.623                         & \colorfirst18.647 & \colorfirst0.658 & \colorsecond0.369 & \colorsecond16.234 & \colorsecond0.600 & \colorfirst0.330 \\\hline
\end{tabular}}
\vspace{-1em}
\end{table*}

\subsection{More Rendering and Mapping Results}
\label{sec:supply:moreresults}
We collect a Livox MID-360 (\textit{private-360}) and a HESAI Pandar XT-32 (\textit{private-pandar}) sequence with image resolution 640$\times$512 to verify the performance of our algorithm in different LiDAR types additionally. The expanded comparision of rendering performance are shown in \mytabref{tab:supply:rendering}.

\begin{figure*}[!htb]
\centering
\includegraphics[width=\linewidth]{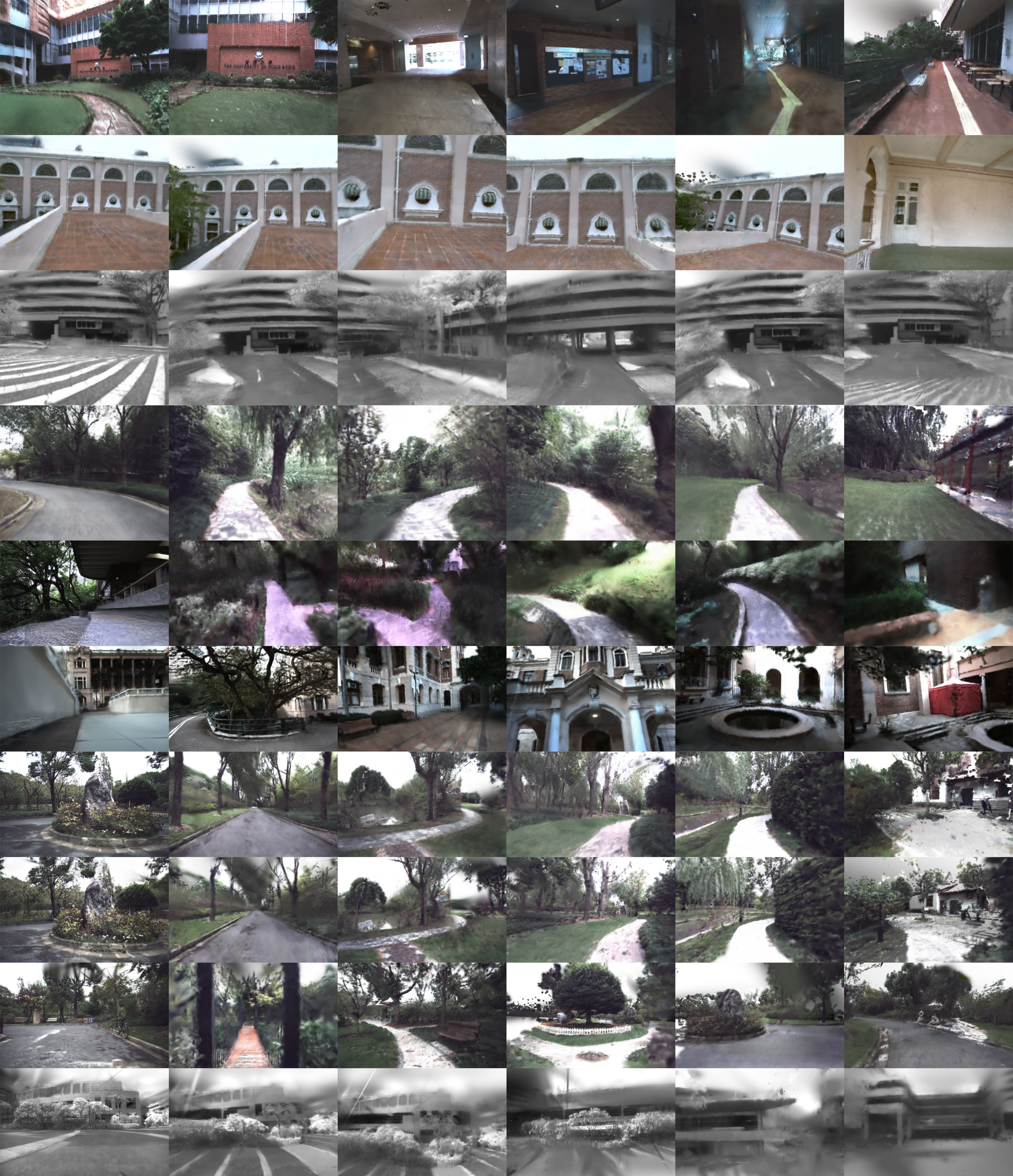}
\caption{Illustration of the rendering performance of the entire sequence from a dataset. It is noteworthy that these illustrations are all rendered based on the real-time mapping results. These scenes are \textit{hkust\_campus\_seq\_00 (Livox AVIA)}, \textit{Visual\_Challenge (Livox AVIA)}, \textit{eee\_03 (Ouster 16)}, \textit{1005\_00 (Livox AVIA)}, \textit{hku\_park\_00 (Livox AVIA)},
\textit{hku\_main\_building (Livox AVIA)}, \textit{1006\_01 (Livox AVIA)}, \textit{1006\_01 (Velodyne VLP-16)}, \textit{1005\_01 (Velodyne VLP-16)}, and \textit{eee\_03 (Ouster 16)}, respectively.
}
\label{fig:supply:render}
\vspace{-1em}
\end{figure*}

It is worth noting that our method can reconstruct the entire scene with limited GPU resources (8GB), which is a feature unattainable by other methods~\cite{hong2024liv, lang2024gaussian}.
In~\myfigref{fig:supply:render}, we present snapshots of rendered images in corresponding whole sequences.
It can be observed that our method maintains excellent rendering quality even in large outdoor scenes under the real-time mapping requirement.

Since our 3D Gaussians information is initialized from point clouds, the rendering performance for sparse multi-line spinning LiDAR is inferior to that of the Livox LiDAR due to the difference in point cloud density. 
This difference can be observed in the Botanic Garden sequence in~\myfigref{fig:supply:render}.

\begin{figure*}[!htb]
\centering
\includegraphics[width=\linewidth]{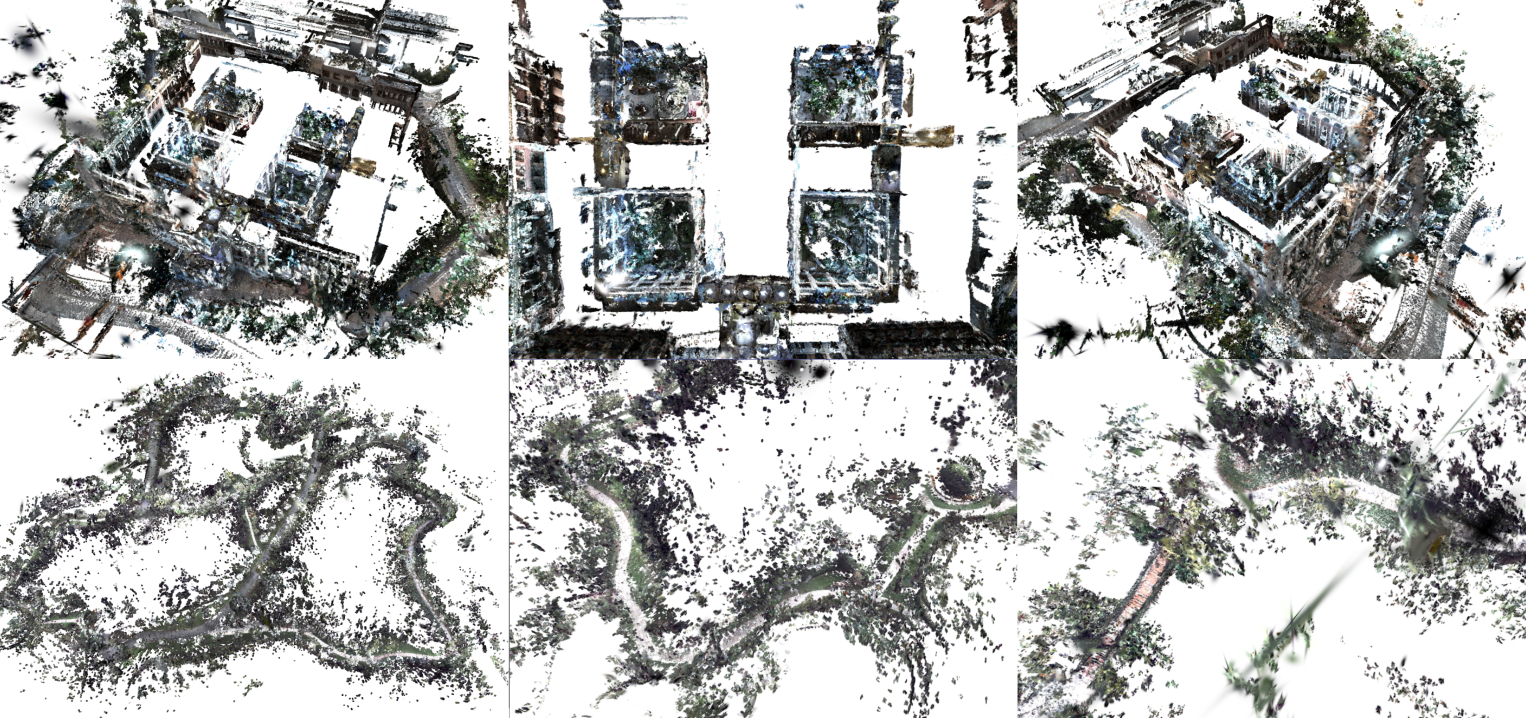}
\caption{More illustrations of the 3D Gaussians model of the entire sequence. These two scenes are \textit{hku\_main\_building ((Livox AVIA))} and \textit{1005\_01 (Velodyne VLP-16)}, respectively.}
\label{fig:supply:model}
\end{figure*}

The reconstructed 3D Gaussian models of sequences \textit{hku\_main\_building} and \textit{1005\_01} are also presented in \myfigref{fig:supply:model}. Sequence \textit{hku\_main\_building} has a duration of 20 minutes and a trajectory length of approximately 900 meters, yet we are still able to achieve real-time map expansion. Outdoor park (sequence \textit{1005\_01}) typically contains rich structural and textural features, making photo-realistic mapping highly challenging.
Using Velodyne VLP-16 LiDAR in the outdoor park, our proposed algorithm is able to achieve a good dense reconstruction model. 

\subsection{More Ablation Experiments}
Expanded ablation experiments on additional sequences are presented in~\mytabref{tab:supply:abla}. In challenging environments, such as sensor degradation (\textit{Visual\_Challenge}) or very sparse point clouds (\textit{eee\_02}), the experiments continue to show that our module design is robust, enhancing photorealistic mapping effects in these scenarios.

\begin{table}[!htb]
\caption{Quantitative ablation comparison of hyperparameters in various sequences, such as duration (DT), mapping time (MT), interval of 3D Gaussians in voxel side $n_s$, convergence variance of Voxel-GPR $\eta$, structure similarity loss (SS), and delta depth similarity loss (DDS). 
}
\centering
\label{tab:supply:abla}
\resizebox{0.9\linewidth}{!}{
\begin{tabular}{@{}ccccccccc@{}}
\toprule
                                                                                             Sequence & DT (s)   &MT (s)                 & $n_s$ & $\eta$ & SS & DDS &  PSNR  & SSIM                          \\ \midrule
                                                                                               &                      & 166  & 3    & 0.1  & \ding{55}    & \ding{55}   & 18.479                         & 0.608                         \\
                                                                                               &                       & 162 & 3    & 0.3  & \ding{55}    & \ding{55}   & 17.326                         & \cellcolor[HTML]{FFD9B3}0.617 \\
                                                                                               &                        & 162 & 3    & 0.3  & \ding{51}    & \ding{55}  & \cellcolor[HTML]{FFD9B3}17.599 & \cellcolor[HTML]{FFF5B3}0.626 \\
                                                                                               &                       & 162 & 3    & 0.3  & \ding{51}    & \ding{51}   & \cellcolor[HTML]{FFF5B3}18.127 & 0.613                         \\
 \multirow{-5}{*}{\textit{\begin{tabular}[c]{@{}c@{}}Visual\_\\ Challenge\end{tabular}}}       & \multirow{-5}{*}{162} & 162 & 4    & 0.3  & \ding{51}    & \ding{51}    & \cellcolor[HTML]{C0E2CA}18.422 & \cellcolor[HTML]{C0E2CA}0.659 \\ \midrule
                                                                                               &                       & 321  &3    & 0.2  & \ding{55}    & \ding{55}  &  12.766                         & 0.436                         \\
                                                                                               &                       &321  & 3    & 0.3  & \ding{55}    & \ding{55}  &  12.136                         & 0.430                         \\
                                                                                               &                       & 321  &3    & 0.3  & \ding{51}    & \ding{55}  &  \cellcolor[HTML]{FFD9B3}12.395 & \cellcolor[HTML]{FFD9B3}0.434 \\
                                                                                               &                       & 321  &3    & 0.3  & \ding{51}    & \ding{51}  &  \cellcolor[HTML]{FFF5B3}12.512 & \cellcolor[HTML]{FFF5B3}0.441 \\
 \multirow{-5}{*}{\textit{eee\_02}}                                                            & \multirow{-5}{*}{321} & 321  &4    & 0.3  & \ding{51}    & \ding{51}  &  \cellcolor[HTML]{C0E2CA}14.328 & \cellcolor[HTML]{C0E2CA}0.473 \\ \bottomrule
\end{tabular}
}
\vspace{-1em}
\end{table}

\begin{table*}[!htb]
\caption{Quantitative tracking performance comparison of LiDAR-IMU SLAM method~\cite{fastlio2} and LiDAR-Visual-IMU fusion SLAM method~\cite{r3live,shan2021lvi} on BotanicGarden dataset~\cite{liu2023botanicgarden}.
}
\label{tab:tracking}
\centering
\resizebox{0.80\linewidth}{!}{
\begin{tabular}{@{}clcccccccllllll@{}}
\toprule
                           & \multicolumn{2}{c}{1005\_00}                                  & \multicolumn{2}{c}{1005\_01}                                  & \multicolumn{2}{c}{1005\_07}                                  & \multicolumn{2}{c}{1006\_01}                                  & \multicolumn{2}{c}{1008\_03}                                  & \multicolumn{2}{c}{1018\_00}                                  & \multicolumn{2}{c}{1018\_13}                                  \\
\multirow{-2}{*}{}         & RPE↓                           & \multicolumn{1}{l}{ATE↓}       & \multicolumn{1}{l}{RPE↓}       & \multicolumn{1}{l}{ATE↓}       & \multicolumn{1}{l}{RPE↓}       & \multicolumn{1}{l}{ATE↓}       & \multicolumn{1}{l}{RPE↓}       & \multicolumn{1}{l}{ATE↓}       & RPE↓                           & ATE↓                           & RPE↓                           & ATE↓                           & RPE↓                           & ATE↓                           \\ \midrule
\multicolumn{1}{l}{R3LIVE~\cite{r3live}} & 1.165                         & \multicolumn{1}{l}{3.153}     & \multicolumn{1}{l}{1.151}     & \multicolumn{1}{l}{1.451}     & \multicolumn{1}{l}{2.112}     & \multicolumn{1}{l}{3.893}     & \multicolumn{1}{l}{\colorthird0.934}     & \multicolumn{1}{l}{3.505}     & 2.050                         & 3.383                         & \colorthird0.165                         & 0.378                         & \colorthird0.133                         & 0.366                         \\
FAST-LIO2~\cite{fastlio2}  & \colorthird1.048 & \colorthird2.665 & \colorthird0.652 & \colorthird0.483 & \colorthird0.947 & \colorthird0.751 & 1.047 & \colorthird1.521 & \colorthird0.852 & \colorthird0.798 & 0.241 & \colorthird0.187 & 0.245 & \colorthird0.308 \\
LVI-SAM~\cite{shan2021lvi}   & \colorsecond0.347 & \colorsecond0.312 & \colorsecond0.147 & \colorsecond0.129 & \colorsecond0.127 & \colorfirst0.257 & \colorsecond0.462 & \colorfirst0.410 & \colorsecond0.272 & \colorfirst0.252 & \colorsecond0.139 & \colorfirst0.044 & \colorsecond0.152 & \colorfirst0.051 \\
OURS                       & \colorfirst0.174 & \colorfirst0.291 & \colorfirst0.058 & \colorfirst0.073 & \colorfirst0.061 & \colorsecond0.496 & \colorfirst0.202 & \colorsecond0.702 & \colorfirst0.068 & \colorsecond0.414 & \colorfirst0.055 & \colorsecond0.075 & \colorfirst0.050 & \colorsecond0.077 \\ \bottomrule
\end{tabular}}
\vspace{-1em}
\end{table*}

\subsection{Tracking Performance}
\label{sec:tracking}
Our photo-realistic mapping method utilizes a general SLAM framework to obtain precise camera poses. In theory, this approach can be integrated into any general LIVO framework. In our experiments, we adopt the framework from~\cite{yuan2024sr}. Compared to the original~\cite{yuan2024sr}, we introduc the following improvements: 1) We replace the time sweep-based refinement method with the sensor's original timestamps at a finer granularity, resolving the crash issue when using multi-line spinning LiDAR. 2) We apply CUDA encapsulation for certain large matrix multiplications to accelerate optimization.
To ensure the fairness of comparisons, we conduct tracking experiments on the improved version of~\cite{yuan2024sr}.

We compare improved method with filter-based method R$^3$LIVE~\cite{r3live}, FAST-LIO2~\cite{fastlio2}, and graph optimization based method LVI-SAM~\cite{shan2021lvi}. In tracking performance evaluation, we focus on key performance metrics Relative Pose Error (RPE) and Absolute Trajectory Error (ATE)~\cite{ate}, considering full transformations that encompass both rotation and translation.

\begin{figure}[!htb]
\centering
\includegraphics[width=\linewidth]{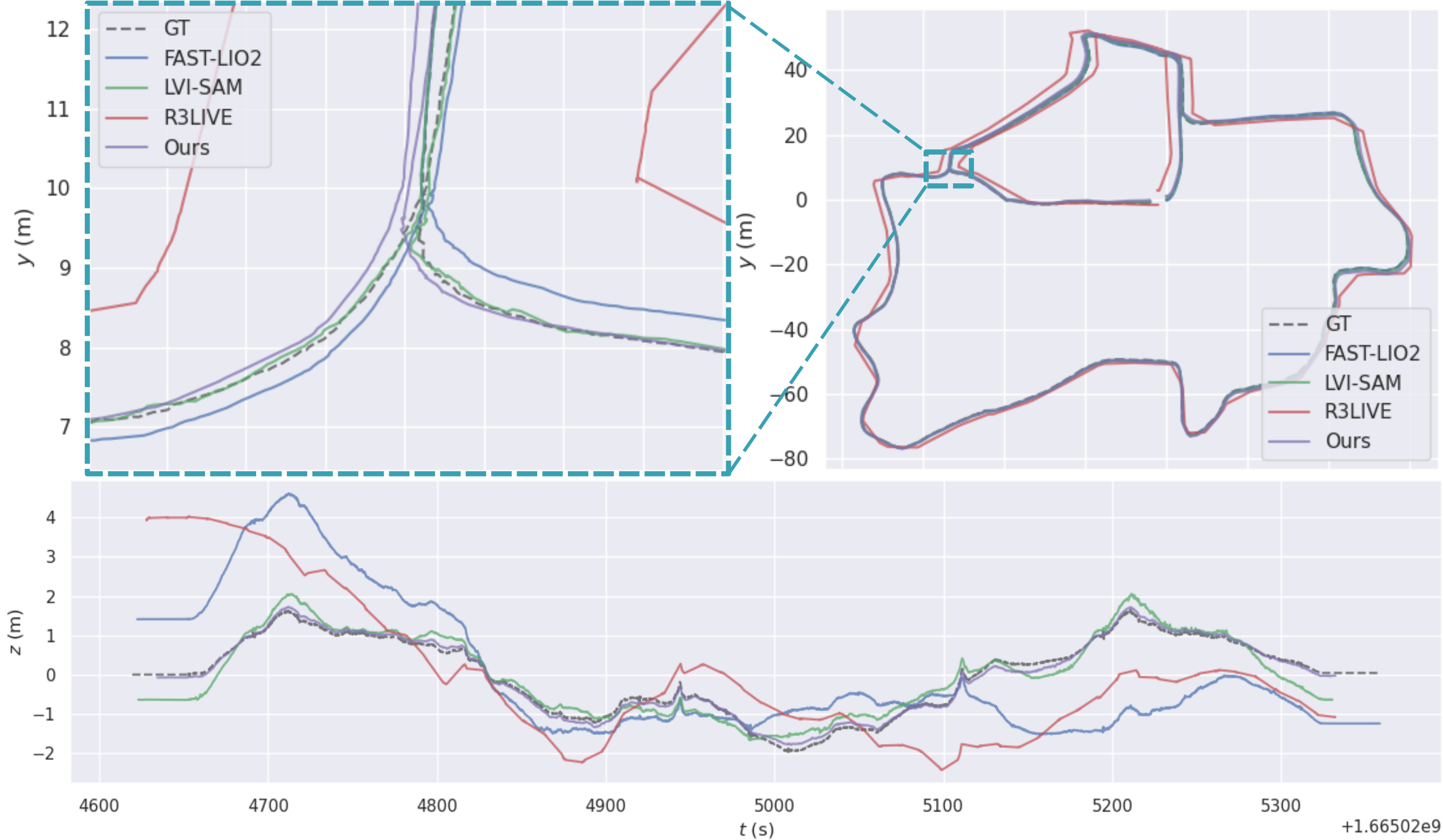}
\caption{The trajectory comparison for the Botanic Garden sequence \textit{1006\_01} is illustrated, featuring a detailed illustration of a \textit{three-way intersection} within the scene in the upper left, with the offset along the \textbf{Z} plane for this junction depicted below. It is evident from this comparison that our trajectory aligns most closely with the actual trajectory, demonstrating a high level of tracking precision.}
\label{fig:tracking}
\vspace{-1em}
\end{figure}

\begin{figure*}[!htb]
\centering
\includegraphics[width=\linewidth]{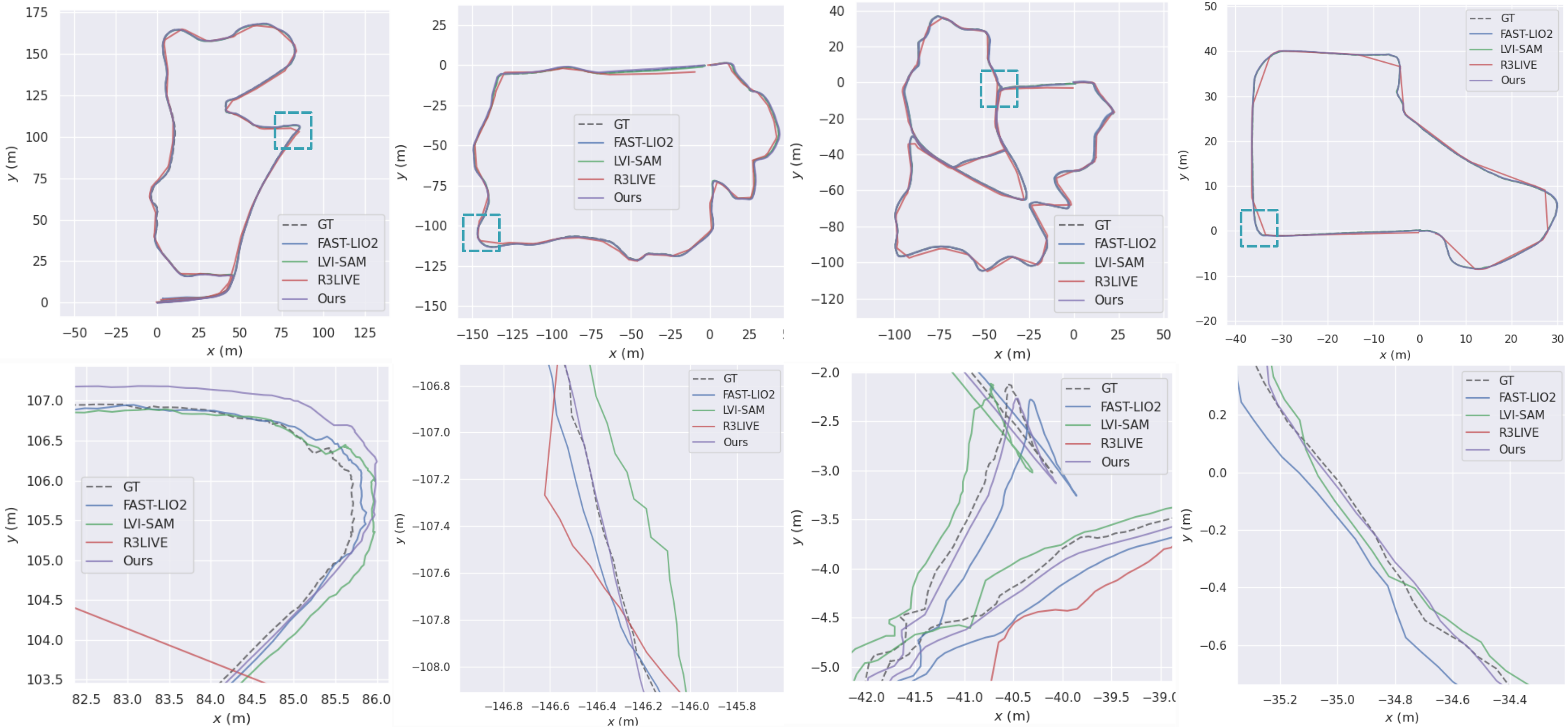}
\caption{More trajectory comparisons. The first row of images shows the trajectory of the entire scene, while the second row of images displays the enlarged results within the corresponding blue boxes. The four scenes are \textit{1005\_00}, \textit{1005\_07}, \textit{1008\_03}, \textit{1008\_13}, respectively.
}
\label{fig:supply:tracking}
\vspace{-1em}
\end{figure*}

To assess the adaptability of our dataset for tracking, we meticulously select 7 representative sequences from the Botanic dataset~\cite{liu2023botanicgarden} and conduct comprehensive evaluations on cutting-edge algorithms, including those presented in~\cite{shan2021lvi, r3live, fastlio2}, by comparing them against ground truth data. 
The data used for our tracking algorithm comes from a Velodyne VLP-16 multi-line spinning LiDAR.
\mytabref{tab:tracking} presents the evaluation metrics on 7 sequences from the Botanic Garden~\cite{liu2023botanicgarden} dataset. Our method, lacking a loop closure detection module, exhibits lower accuracy on some sequences compared to LVI-SAM~\cite{shan2021lvi}. Overall, comparisons with above state-of-the-art methods reveal that our approach demonstrates clear advantages on certain sequences. ~\myfigref{fig:tracking} illustrates the trajectory comparison for the sequence \textit{1006\_01} within the Botanic Garden~\cite{liu2023botanicgarden} dataset, featuring a trajectory length of approximately 730 meters. In the top right corner of ~\myfigref{fig:tracking}, an enlarged view of a \textit{three-way junction} on the \textbf{X-Y} plane is presented, while the bottom right corner displays an elevation map of the same junction on the \textbf{Z} plane. From the visual representation, it is evident that our trajectory aligns most closely with the ground truth, showcasing the highest qualitative tracking precision. 
More comparison results are available on \myfigref{fig:supply:tracking}.

\begin{table}[!htb]
\caption{Duration (DT), mapping time (MT), count of 3D Gaussians, maximum memory cost (Mem), and rendering metrics on ultra long sequence.
}
\label{tab::supply::dura-gs-gpu}
\resizebox{\linewidth}{!}{
\begin{tabular}{@{}ccccccc@{}}
\toprule
Sequence                 & DT (s) & MT (s)   & Count     & Mem (Mb)  & PSNR   & SSIM  \\ \midrule
\textit{hku\_main\_building} & 1170     & 1170 & 2,674,234 & 4489 & 15.234 & 0.538 \\
\textit{hkust\_campus\_00} & 1073     & 1090 & 3,302,675 & 5812 & 15.124 & 0.496\\ \bottomrule
\end{tabular}}
\end{table}

\begin{figure}[!htb]
\centering
\includegraphics[width=\linewidth]{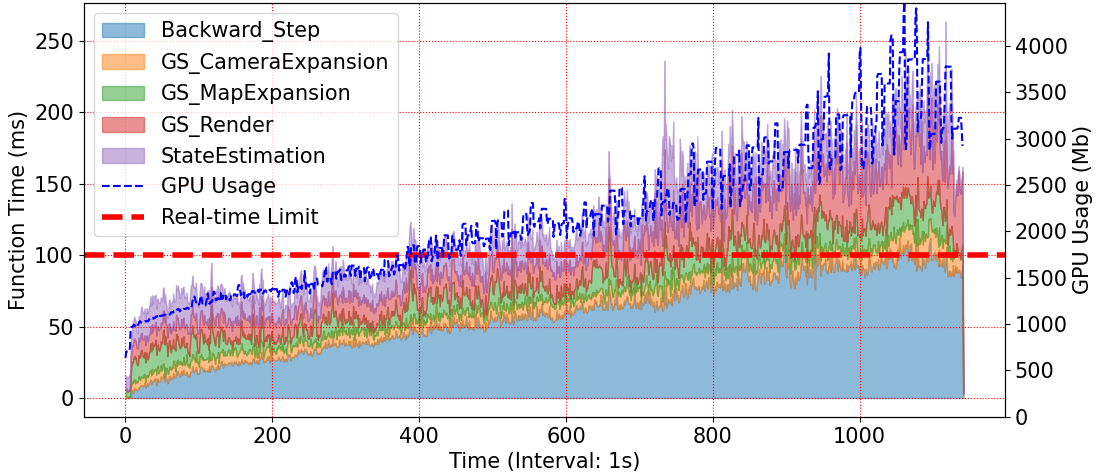}
\caption{Time consumption and GPU memory consumption on ultra long scequence \textit{hku\_main\_building} (Duration: 1170\,s).}
\label{fig::supply::timegpu}
\end{figure}

\begin{table}[!htb]
\caption{Hyper parameters used in all datasets.}
\resizebox{\linewidth}{!}{
\begin{tabular}{@{}ccc@{}}
\toprule
 Name                             & Value        & Description                                                   \\ \midrule
 size                             & 0.2 & Side length of each voxel.                              \\
 $\tau$                           & 10           & Point count threshold for voxel processing.  \\
 $n_s$                            & 3            & Interval between each side within the voxel.                   \\
 $n_r$                            & 3            & Number of neighboring Gaussians per side.  \\
 $\eta$                           & 0.3          & Convergence threshold for Voxel-GPR.                        \\ \midrule
 $\lambda_{SSIM}$                  & 0.2          & Weight for the photometric SSIM loss.                             \\
 $\lambda_p$                       & 0.1          & Weight for the structural similarity loss.                         \\
 $\lambda_d$                       & 0.1          & Weight for delta depth similarity loss.                       \\
 $\mathcal{T}$                     & 50           & Size of the sliding window for recent images.                              \\
 $k_{curr}$                        & 1            & Number of frames selected from the current window.        \\
 $k_{his}$                         & 1            & Number of frames selected from historical data.            \\ \midrule
 $lr_{position}$                   & 0.0005       & Learning rate for position parameters.            \\
 $lr_{color}$                      & 0.0025       & Learning rate for color parameters.               \\
 $lr_{opacity}$                    & 0.025        & Learning rate for opacity parameters.             \\
 $lr_{scale}$                      & 0.0025       & Learning rate for scale parameters.    \\
 $lr_{rotation}$                   & 0.0025       & Learning rate for rotation parameters. \\ \bottomrule
\end{tabular}
}
\label{tab:hyperparameters}
\vspace{-1em}
\end{table}

\subsection{Time Consumption and Memory Usage on Ultra-Long Time Sequence}
\label{sec:supply:timegpu}
Addressing SLAM and photo-realistic mapping in ultra-long time sequence is a challenging issue. As the scale of the map increases, insufficient available storage space leads to a slowdown or interruption in the mapping process. To deal with these problems, we conduct tests on two ultra-long sequences in test dataset. With the limited GPU resources, we set $n_r$=2 and obtain the mapping time, rendering metrics, as shown in \mytabref{tab::supply::dura-gs-gpu}. The time and GPU memory consumption of each algorithm module over time are depicted in \myfigref{fig::supply::timegpu}. In the later stages of the long time series, most of the time is spent on rendering the ultra-large 3DGS map, map feedback, and optimization. It can be seen that our method is still able to achieve real-time map expansion in ultra-long sequences, with resource consumption in ultra-large scenes remaining within acceptable levels.

\subsection{Hyper Parameters in Our Experiments}
The hyper parameters used in all our sequences are listed in \mytabref{tab:hyperparameters}. For more configuration details, please refer to the \textit{config} file in our GitHub repository.

\section{Limitations}
In this paper, real-time performance is our primary focus, with all modules specifically designed to minimize the time complexity of the algorithm, thus ultimately enabling real-time photo-realistic mapping in large-scale open environments. However, the study does have some limitations:
The initialization of the 3DGS model is solely reliant on the point cloud data, which results in initialization failure in areas lacking point cloud coverage, leading to missing regions. Future research will explore integrating image data to aid in the initialization process.
Moreover, due to the real-time constraints, the reconstruction quality may not match that of images obtained through actual data acquisition. Improving the reconstruction quality will be part of our future research endeavors.
\end{document}